\documentclass{article}

\usepackage{arxiv}
\usepackage{xcolor}
\usepackage{amsfonts}
\usepackage{ulem}
\usepackage{natbib}
\usepackage{graphicx}
\usepackage[justification=centering]{caption}
\usepackage[justification=centering]{subcaption}
\usepackage{url}

\usepackage{siunitx}

\title{Combining distribution-based neural networks to predict weather forecast probabilities}

\author{
  Mariana C A Clare \\
  Imperial College London,\\
  London, UK\\
  \texttt{m.clare17@imperial.ac.uk} \\
   \And
Omar Jamil \\
University of Exeter,\\
Exeter, UK \\
(\textit{formerly Met Office, Exeter, UK})\\
\AND
Cyril Morcrette\\
Met Office,\\
Exeter, UK\\}

\begin{document}

\maketitle

\begin{abstract}
		The success of deep learning techniques over the last decades has opened up a new avenue of research for weather forecasting. Here, we take the novel approach of using a neural network to predict full probability density functions at each point in space and time rather than a single output value, thus producing a probabilistic weather forecast.  This enables the calculation of both uncertainty and skill metrics for the neural network predictions, and overcomes the common difficulty of inferring uncertainty from these predictions.
		
		This approach is data-driven and the neural network is trained on the WeatherBench dataset (processed ERA5 data) to forecast geopotential and temperature 3 and 5 days ahead.  Data exploration leads to the identification of the most important input variables, which are also found to agree with physical reasoning, thereby validating our approach. In order to increase computational efficiency further, each neural network is trained on a small subset of these variables. The outputs are then combined through a stacked neural network, the first time such a technique has been applied to weather data. Our approach is found to be more accurate than some numerical weather prediction models and as accurate as more complex alternative neural networks, with the added benefit of providing key probabilistic information necessary for making informed weather forecasts.
\end{abstract}


	\section{Introduction}\label{sec:intro}
	
	For over 100 years, advanced mathematical techniques have been used for weather prediction. Today, Numerical Weather Prediction (NWP) is an advanced discipline which uses some of the world's largest supercomputers to solve complex non-linear differential equations. The forecast skill of these models has been improving by approximately one day every ten years, \textit{i.e.} the 5-day forecast today is as accurate as the 4-day forecast was ten years ago \cite[see][]{Bauer2015}. This improvement has been achieved through the scientific and technological development of both NWP models and computers \cite[]{Bauer2015}. However, the success of deep learning techniques over the last decade has opened up a new avenue for weather forecasting \cite[]{Schultz2021}. Research has been mainly focused on the supervised learning techniques of neural networks \cite[]{Dueben2018, Brenowitz2019, Rasp2020, Rasp2020a, Weyn2020} and random forests \cite[]{Yuval2020}. Some works have combined NWP models with neural networks: for example \cite{Brenowitz2019} successfully couple neural networks with General Circulation Models to emulate physical parametrisations and \cite{rasp2018neural} and \cite{gronquist2021deep} use neural networks to post-process ensemble weather forecasts from NWP models. Other works have taken a purely data driven approach \cite[e.g.][]{Rasp2020a}. In this work, we take the purely data-driven approach and use residual neural networks to create 3-day and 5-day hindcasts.
	
	A limitation of the deep learning approaches used in the works referred to above is that it is difficult to infer the uncertainty of the predictions from their results \cite[]{Schultz2021}. Some previous works address these limitations by using an ensemble of deep learning models to produce a probabilistic forecast \cite[e.g.][]{Scher2020, Bihlo2021, Weyn2021}. However, choosing a good ensemble of models is non-trivial (see \cite{Scher2020}) and may be computationally expensive because it requires the network to be trained multiple times. Others have dealt with the issue of uncertainty by training neural networks to predict from the weather data itself, the error and ensemble spread which would be produced if an NWP model were applied to this data \textit{i.e.} a mixed data-driven neural network and NWP approach \citep[]{scher2018predicting}. This method is not purely data driven though and requires access to good NWP forecasts. 
	
	In this work, we propose a novel approach to deal with the issue of assessing uncertainty from neural network outputs. With our approach, the neural networks predict full probability density functions for the target variable at each point in space and time instead of single values. These density functions allow practitioners to estimate the uncertainty of the neural network outputs and make a more informed weather forecast. In order to reduce computational cost and optimise model accuracy, we train multiple neural networks on a small number of variables and combine their outputs using techniques such as a stacked neural network. This is a technique which has not been used with weather data before. In this work, the neural networks are trained on the WeatherBench dataset created by \cite{Rasp2020} and used to predict both a 3-day and a 5-day weather hindcast of geopotential at the 500hPa pressure level in \SI{}{m^{2}s^{-2}} (hereafter Z500) and temperature at the 850hPa pressure level in Kelvin (K) (hereafter T850). These variables are chosen so that our results can be compared to those in other works which use the same dataset \cite[]{Rasp2020, Rasp2020a, Weyn2020}. 
	
	The remainder of this work is structured as follows: Section \ref{sec:methods} describes the data used in this study followed by the neural network architectures and data exploration techniques used; Section \ref{sec:results} presents results from using stacked neural networks to forecast weather data and shows how the output can be used to infer uncertainty; and finally Section \ref{sec:conclusion} concludes the work.
	
	\section{Methods}\label{sec:methods}
	\subsection{Data}\label{subsec:data}
	The WeatherBench dataset is a global dataset produced by \cite{Rasp2020} containing a mix of multi-levelled (13 pressure levels) and single level variables. It uses as its raw data the ERA5 reanalysis dataset \cite[]{hersbach2020era5} for the 40-year period from 1979 to 2018. The data was processed and regridded onto a $5.625^{o}$ resolution latitude-longitude grid (32 $\times$ 64 grid points) by \cite{Rasp2020} and we refer the reader to that work for more details. Following the same work, we consider data from 2017 to 2018 to be the test dataset. One of the benefits of deep learning is that we do not need to carry out extensive feature engineering and the neural networks are able to find the best predictors in the data. However, it is still necessary, as a first step, to choose an appropriate architecture for the neural network. For this first step, we use data from 1979 to 2015 as the training dataset with data from 2015 used for validation of the neural network (hereafter referred to as the neural-validation dataset). All the results in this section are applied on the 2016 data (hereafter referred to as the validation dataset), so that results on the test dataset are not used to make any architecture decisions.
	
	\subsection{Neural network architectures}\label{subsec:neural_network}
\begin{figure}[ht]
		\centering
		\includegraphics[width = \textwidth]{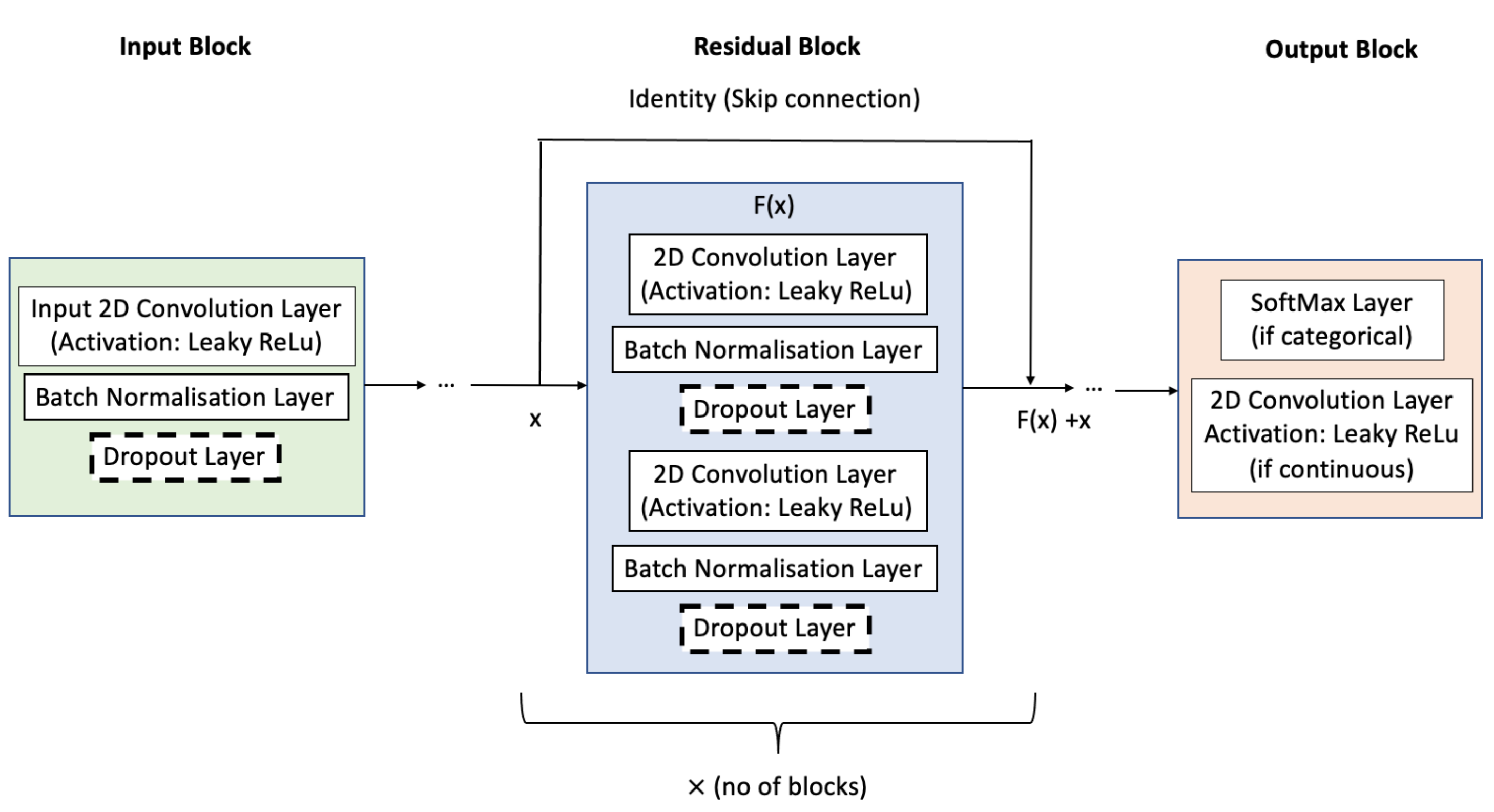}
		\caption[Schematic of convolutional ResNet]{Schematic of convolutional ResNet used in this work. Each 2D convolution layer has LeakyReLu activation with $\alpha = 0.3$ \citep{maas2013} and 100 channels (because of 100 bins) if categorical data is being trained or 64 channels if continuous data is being trained. Following \cite{Rasp2020}, the 2D convolutions are defined with periodic padding in the longitude direction and zero padding in the latitude direction with a kernel size of 5. The dropout layer has a dropout rate of 0.1. If dropout is not required then there is no dropout layer in the network architecture. Note, each ResNet takes between approximately 6 and 12 hours to train (depending on the size of the input data and the number of residual blocks) on a RTX6000 machine with two GPUs and 48GB of memory.}\label{resnet}
	\end{figure}

	Fundamentally, a neural network provides a way to extract non-linear relationships present in the data and is trained to minimise a loss. This minimisation is done via gradient descent which is used to update the neural network weights. Previous works have applied several different types of neural networks to this dataset: the original WeatherBench dataset work \cite{Rasp2020} uses a simple convolutional neural network (CNN), \cite{Weyn2020} uses a U-Net (an architecture first described in \cite{u_net}), and \cite{Rasp2020a} uses a 19-block convolutional ResNet (an architecture first described in \cite{he2016deep}). The latter results in the lowest errors of them all and thus in this work we choose this architecture. 
	Generally, residual neural networks consist of a series of repeated blocks (referred to as residual blocks) of convolutional, normalisation, and dropout layers, with intermediate connections known as `skip connections'. A skip connection between residual blocks adds the outputs from previous blocks to the output of the current block (see Figure \ref{resnet}). In this way they can avoid the issue of accuracy saturation which occurs in other types of neural network architectures when more layers are added \citep{he2016deep}. Figure \ref{resnet} shows the general structure of the convolutional ResNet used in this work and provides details of the layers used. It highlights that, as discussed in \cite{lu2018}, a ResNet can be viewed as a variation of the forward Euler finite difference method 
	\begin{equation}
	y_{n+1} = y_{n} + hf(t_{n}, y_{n}),
	\end{equation}
	which is actually a much simpler version of the methods used by NWP models to predict the weather \cite[]{White1971}. This makes a ResNet an appropriate architecture for predicting weather and partly explains the greater accuracy given by the convolutional ResNet in \cite{Rasp2020a} relative to other neural networks.
	
\begin{figure}[ht]
		\begin{subfigure}{0.48\textwidth}
			\centering
			\includegraphics[width =0.9\textwidth]{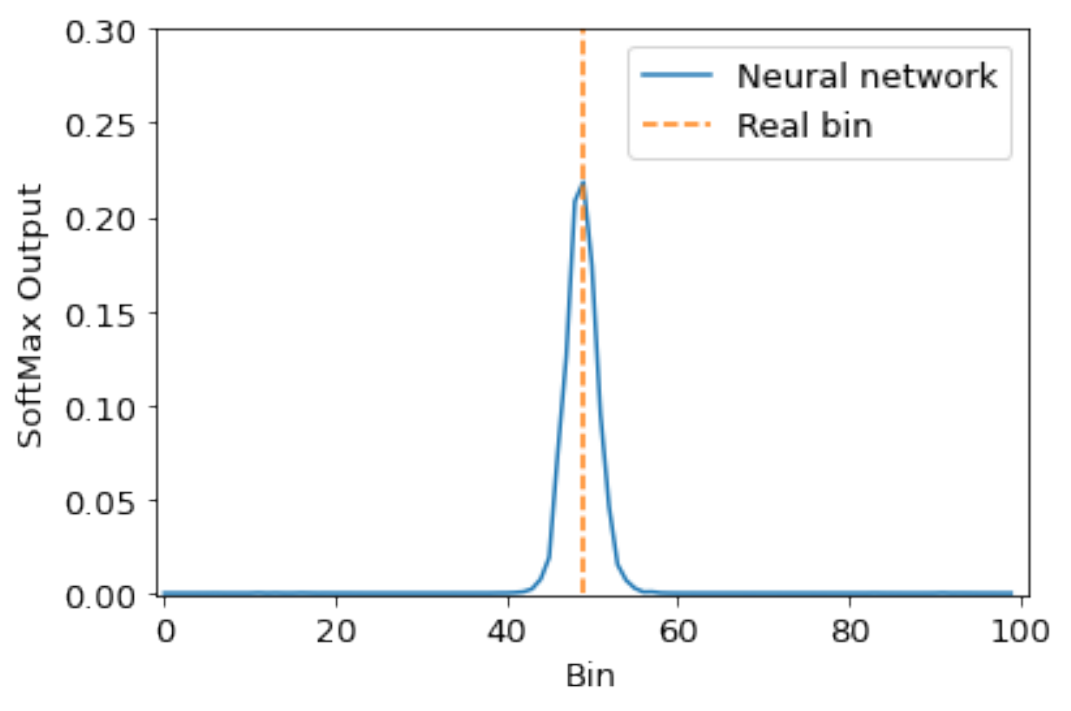}
			\caption{}\label{fig:prob_dist_a}
		\end{subfigure}
		\hfill
		\begin{subfigure}{0.48\textwidth}
			\centering
			\includegraphics[width =0.9\textwidth]{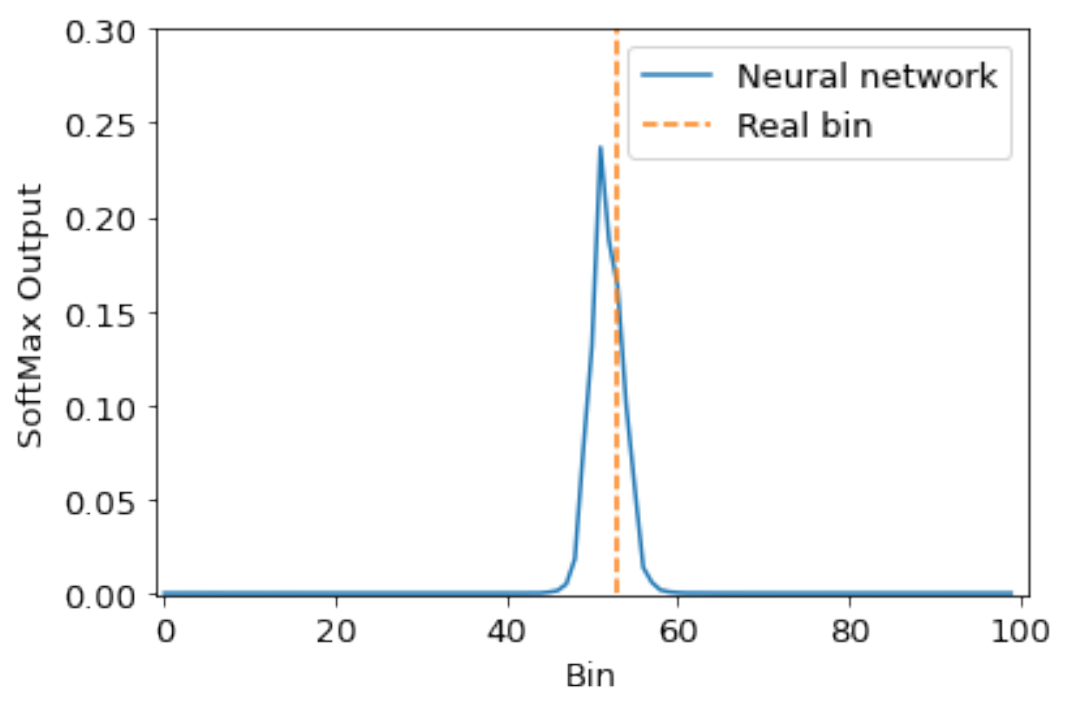}
			\caption{}\label{fig:prob_dist_b}
		\end{subfigure}
		\caption{Two randomly selected examples of the probability density function for the Z500 3-day hindcast at different gridpoints and times. These have been predicted by a ResNet with a SoftMax output layer and 15 residual blocks. Each bin corresponds to a geopotential range of width \SI{169}{m^{2}s^{-2}}, where the lower bound of the 0-bin is \SI{42500}{m^{2}s^{-2}}  and the upper bound of the 99-bin is \SI{59300}{m^{2}s^{-2}}.}
		\label{fig:prob_dist}
	\end{figure}

	The main novelty of our work is that our neural network architecture aims to predict the probability distribution of the variables Z500 or T850 at a particular point in time, longitude and latitude, rather than their exact value which is what the previous works discussed above predicted. In order to achieve this, the first step is to convert the continuous weather data to categorical data by taking each target variable and binning its values into 100 bins of equal width. The values of Z500 vary between a minimum of \SI{42500}{m^{2}s^{-2}} and a maximum of \SI{59300}{m^{2}s^{-2}}, meaning that each Z500 bin has a width of \SI{169}{m^{2}s^{-2}} (to 3 significant figures); T850 varies between \SI{213}{\kelvin} and \SI{314}{\kelvin}, leading to a bin width of \SI{1.02}{\kelvin} (to 3 significant figures). We take the value of the category to be its lowest value, which introduces an inbuilt root mean square error (RMSE) of \SI{91.2}{m^{2}s^{-2}} for Z500 data and \SI{0.992}{\kelvin} for T850 data (calculated using (\ref{RMSE})). With the target variables binned, it is then possible to use a SoftMax layer as the output layer (see Figure \ref{resnet}). The SoftMax layer predicts the probability density functions of the variables by using an activation function which exponentiates the input and then normalises it,
	thus outputting a vector with a value between 0 and 1 for each bin \citep{softmax}. The sum of this vector is equal to 1 meaning it is a probability density. Figure \ref{fig:prob_dist} shows two randomly chosen examples of the output from the SoftMax layer for a categorical output variable. In Figure \ref{fig:prob_dist_a}, the maximum probability bin is the true bin. Although in Figure \ref{fig:prob_dist_b} the maximum probability bin (23.7\%) is not the true bin, the true bin has a non-zero probability of 16.3\%, meaning this probability density function is a useful tool from a weather-forecasting perspective.
	
	Our use of categorical data also necessitates a different choice of loss metric to the previous works, using the WeatherBench dataset \cite[e.g.][]{Rasp2020a}, which use the mean squared error. A loss metric is used by the neural network to calculate the loss during training and it can play a pivotal role in the accuracy and efficiency of a neural network. We choose sparse categorical cross-entropy (from keras) in our neural network due to our use of categorical data and the memory efficiency of this metric. Note, these loss metrics can also be used during the training of the neural network to determine when to stop training and set the values of important parameters. For example, in our case, we compile our neural network with an Adam Optimiser \cite[][]{kingma2014adam} with an initial learning rate of 5e-5. This learning rate is reduced by a factor of 5 if the loss metric on the neural-validation dataset does not decrease after 2 epochs (\textit{i.e.} ater the entire training dataset has passed through the neural network twice). If the loss metric on the neural-validation dataset does not decrease after 5 epochs then the neural network stops training. In general, the neural network requires approximately 15 epochs but this varies depending on the set of inputs being trained.
	
	In addition, to a new loss metric, in order to be able to compare the results of our new categorical data approach with the correct values from the data and with other neural network and numerical model results, it is necessary to infer a single value from our categorical neural network predictions. To do this, we again take advantage of the probability density function we have predicted and calculate their expectation using 
	\begin{equation}\label{expectation}
	\mathbb{E}[X] = \sum_{i = 1}^{100} x_{i}\mathbb{P}(X = x_{i}),
	\end{equation}
	where $x_{i}$ is chosen to be the lower bound of each bin. In this way, we take advantage of the density functions we have predicted and also reduce the inbuilt error caused by binning the data, which we discussed previously. Throughout this work, the RMSE is calculated between the real data and the expected values of the density functions generated by our approach using the latitude-weighted RMSE outlined in \cite[e.g.][]{Rasp2020}. This is given by
	
	\begin{equation}\label{RMSE}
	\text{RMSE} = \sqrt{\frac{1}{N_{\text{timepoints}}} \sum_{i=1}^{N_{\text{timepoints}}}\frac{1}{N_{\text{lat}}N_{\text{lon}}}\sum_{j=1}^{N_\text{lat}}\sum_{k=1}^{N_\text{lon}}L(j)(f_{i,j,k}-t_{i,j,k})^{2}},
	\end{equation}
	where 
	\begin{equation}\label{Lij}
	L(j) = \frac{\cos(\text{lat}(j))}{\frac{1}{N_{\text{lat}}}\sum_{j=0}^{N^{\text{lat}}}\cos(\text{lat}(j))},
	\end{equation}
	is the latitude weighting factor, $f$ is the predicted value from the neural network and $t$ is the true value from the dataset. 
	
	Despite this expectation calculation, there will still be differences caused by using categorical output data compared to continuous data. The simplest way to understand these is by training a series of continuous data and categorical data neural networks with varying numbers of residual blocks and comparing the results. Note that to reduce both the computational cost of training the network and the memory cost, we train two separate networks to predict Z500 and T850 individually. If we had trained a single neural network to predict both Z500 and T850 at the same time, this would have added an extra dimension to the output making it 5-dimensional. We note that, unless explicitly stated, all results shown in this work refer to the 3-day hindcast.
	
	For our initial analysis, we use a training dataset consisting of only the two variables of interest, Z500 and T850 at current time $t$. Figure \ref{resnet} and its caption show the neural network architectures used for training continuous data (the same as that in \cite{Rasp2020a}) and categorical data. Note that, for simplicity, initially we have chosen not to include dropout in either neural network type. 
	
	\begin{figure}[ht]
		\begin{subfigure}{0.49\textwidth}
			\centering
			\includegraphics[width =0.9\textwidth]{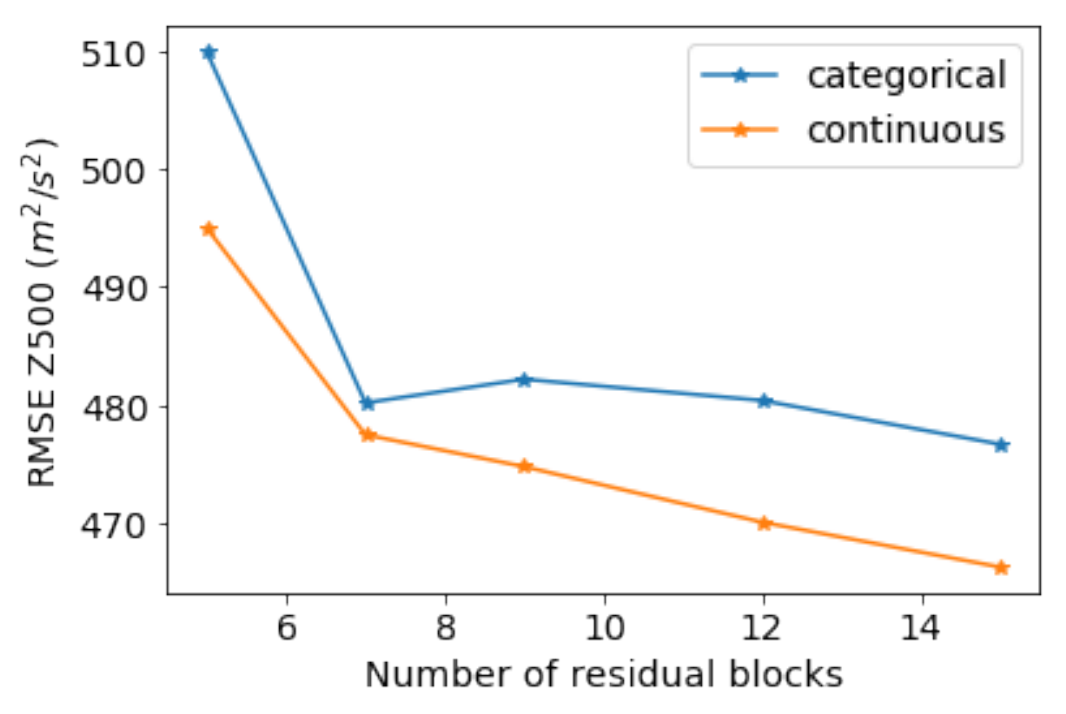}
			\caption{RMSE for 500hPa geopotential.}\label{categorical_z500}
		\end{subfigure}
		\begin{subfigure}{0.49\textwidth}
			\centering
			\includegraphics[width =0.9\textwidth]{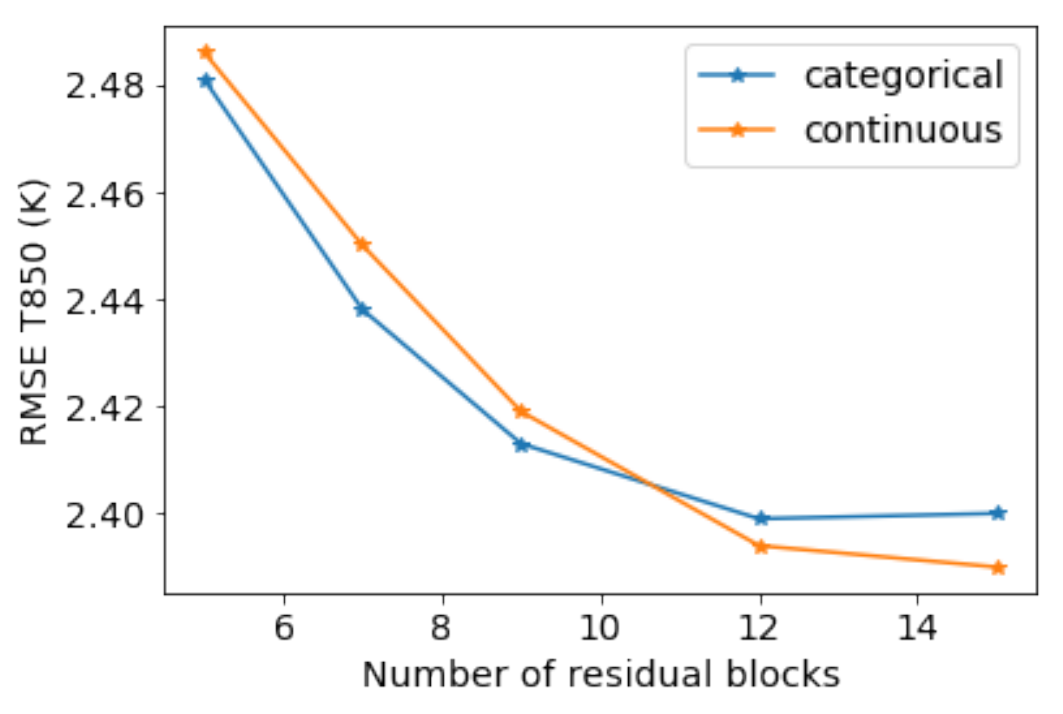}
			\caption{RMSE for 850hPa temperature.}\label{categorical_t850}
		\end{subfigure}
		\caption[Comparison of RMSE]{Comparison of the RMSE achieved by training a neural network with categorical data and that achieved by training a neural network with continuous data. The result is shown for different numbers of residual blocks. The RMSE is a weighted average calculated using (\ref{RMSE}) over all gridpoints and times in the validation dataset.}\label{categorical}
	\end{figure}
	
	Figure \ref{categorical} shows that the errors decrease as the number of residual blocks increases for both categorical and continuous data. For Z500, Figure \ref{categorical_z500} shows that the error from training on continuous data is always less than that from training on categorical data, but the difference between the two errors is much less than the inbuilt binning error of \SI{91.2}{m^{2}s^{-2}}. For T850, Figure \ref{categorical_t850} shows that there is little difference in the error as a result of training on categorical data compared to the error from training on continuous data and in fact for less than 12 residual blocks the categorical data error is lower. This is despite the fact that the inbuilt binning error calculated previously is \SI{0.992}{\kelvin}. This suggests that although working in terms of binned data introduces an error, the new neural network structure and the expectation calculation is able to partially compensate for this.
	
	\subsubsection{Using dropout for ensemble modelling}\label{subsubsec:dropout}
	So far the neural network architectures used in our tests have not included a dropout layer. However, including a dropout layer is a common strategy to improve the performance of neural networks \cite[e.g.][]{srivastava2014dropout}. This layer randomly ignores some outputs from the preceding layer in the network at a rate set by the user (we use a rate of 0.1), meaning these outputs are not passed on to the proceeding layer of the network. This means that if dropout is occurring, the neural network is slightly different every time the data passes through it, which helps prevent overfitting during training. Moreover, if dropout is allowed to occur at the inference/prediction phase then an ensemble of outputs can be generated from a single neural network train \cite[]{Gal2016}. Thus in this section, we examine the improvements that can be achieved from using dropout in our neural network.
	
	Dropout ensemble technique have already been applied on continuous weather data in \cite{Scher2020}, where they show that using this technique results in an improvement in accuracy. With our categorical data approach, each ensemble member is actually a series of density functions at every point in space and time (recall Figure \ref{fig:prob_dist}) and so careful analysis is required before they are combined. The law of total probability states that
	\begin{equation}\label{eq:tot_prob}
	\mathbb{P}(A) = \sum_{n}\mathbb{P}(A|B_{i})\mathbb{P}(B_{i}),
	\end{equation}
	if $B_{i}$ are a finite number of pairwise disjoint sets, whose union is the sample space. In addition, with this technique there is no reason why one ensemble member should be more accurate than the others, and thus a simple average is sufficient to combine them. If we set $\mathbb{P}(A|B_{i})$ to be the probability density function from the $i$th ensemble member and $\mathbb{P}(B_{i})$ to be the probability of sampling from the $i$th distribution (thus $\mathbb{P}(B_{i}) = 1/n$ where $n$ is total number of ensemble members), we show that averaging the density functions from the ensemble members is mathematically rigorous. This averaging is known as linear pooling \cite[]{allard2012probability}.
	
\begin{figure}[ht]
		\begin{subfigure}{0.49\textwidth}
			\centering
			\includegraphics[width =0.9\textwidth]{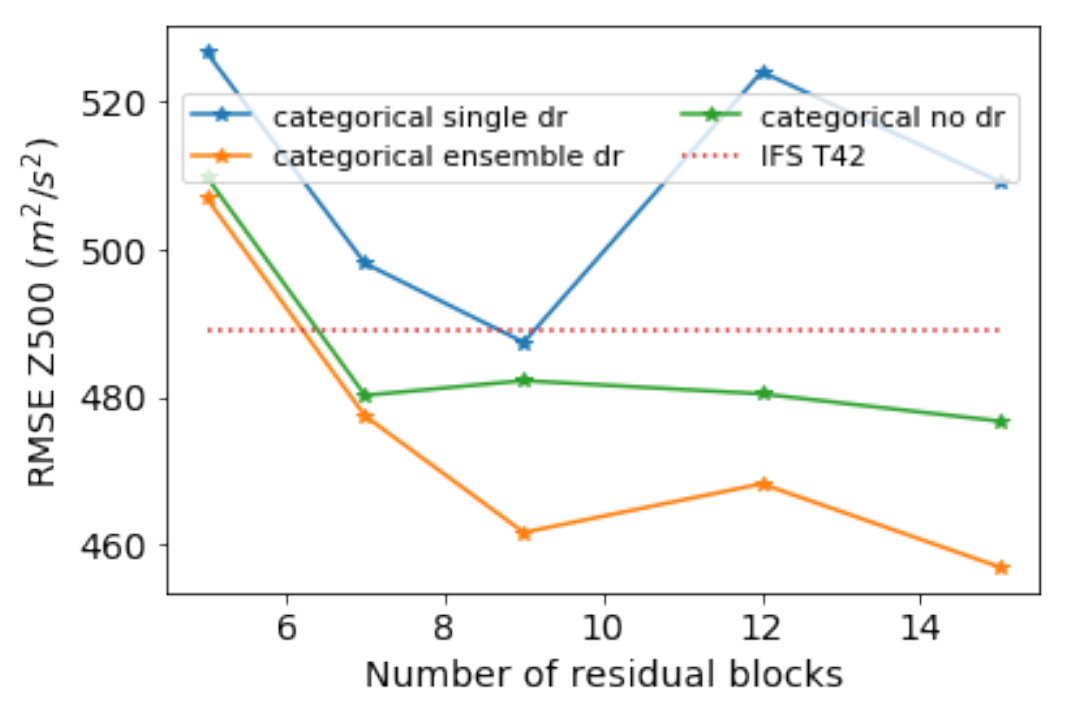}
			\caption{Error for 500hPa geopotential.}
		\end{subfigure}
		\begin{subfigure}{0.49\textwidth}
			\centering
			\includegraphics[width =0.9\textwidth]{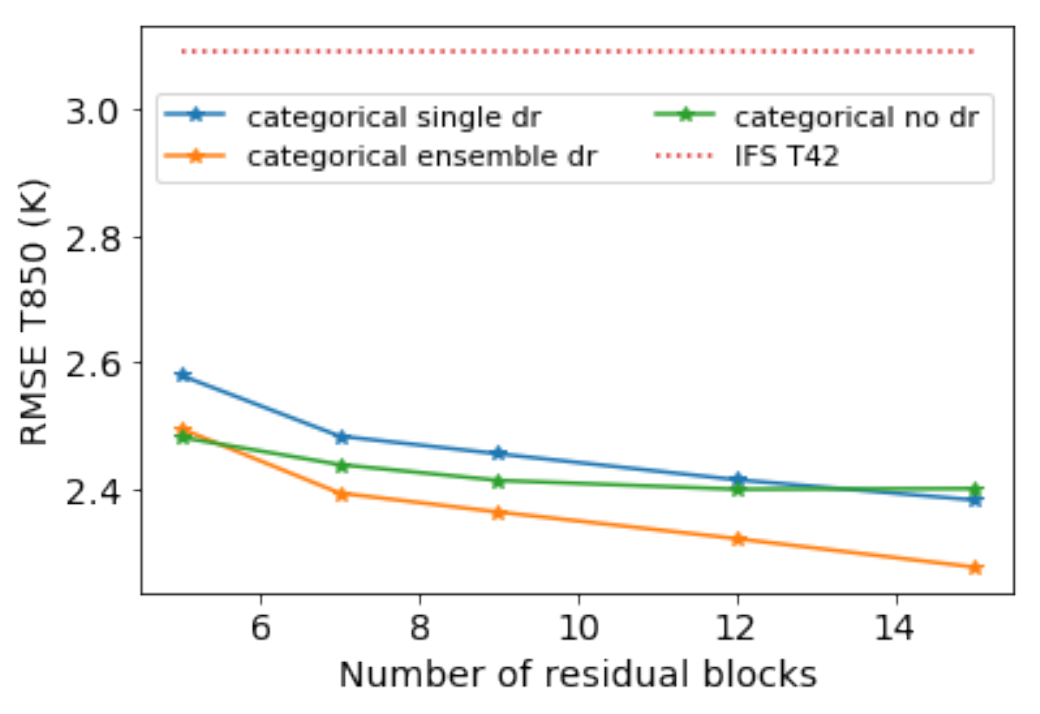}
			\caption{Error for 850hPa temperature.}
		\end{subfigure}
		\caption[RMSE Comparison]{Comparison of the RMSE achieved by using the dropout ensemble technique with 32 ensemble members; using a single output from a neural network with dropout; using a neural network with no dropout and using the coarse NWP model IFS T42. The ensemble dropout technique almost always outperforms the other techniques. Note the RMSE of the neural network is calculated using the weighted average (\ref{RMSE}) over all gridpoints and times in the validation dataset (\textit{i.e.} 2016), whereas the IFS T42 RMSE is calculated using the same method but over the test dataset (\textit{i.e.} 2017-18).}\label{comp_do}
	\end{figure}

	Using this knowledge, we can average the ensemble outputs. Figure \ref{comp_do} shows the results of averaging an ensemble of 32 members and shows that using the dropout ensemble technique results in a notable reduction in error compared to not using dropout at all or using dropout but with only one ensemble member. This reduction in error as a result of the dropout ensemble technique is obtained for almost no extra computational cost and occurs no matter the number of residual blocks used. Whilst for T850 the error always decreases as the number of residual blocks increases, for Z500 the trend is less clear. This may be due to overfitting and the optimum number of residual blocks will be discussed in more detail in Section \ref{sec:results}. For reference, Figure \ref{comp_do} also shows the error from using the configuration of the Integrated Forecast System (IFS) model of the European Centre for Medium-range Weather Forecasting (ECMWF) at the T42 resolution (approximately $2.8^{o}$ resolution at the equator) [IFS T42]. The T42 resolution is much coarser than the operational IFS used by ECMWF, but twice as fine as the resolution of the WeatherBench dataset ($5.625^{o}$) used in this work. Despite this finer resolution, Figure \ref{comp_do} shows that our ResNet with categorical data outputs trained on just Z500 and T850 is substantially more accurate than IFS T42 for both Z500 and T850 when the dropout ensemble technique is used with more than 5 residual blocks. Note this IFS T42 value is for the 2017-2018 data and is taken from \cite{Rasp2020a}, whereas the neural network errors are for the 2016 data, but we have also used the neural networks to predict the 2017-2018 data and found the same results. We have not included this here so as not to mislead the reader by showing preliminary neural network results applied to test data.
	
\begin{figure}[ht]
		\centering
		\includegraphics[width =0.45\textwidth]{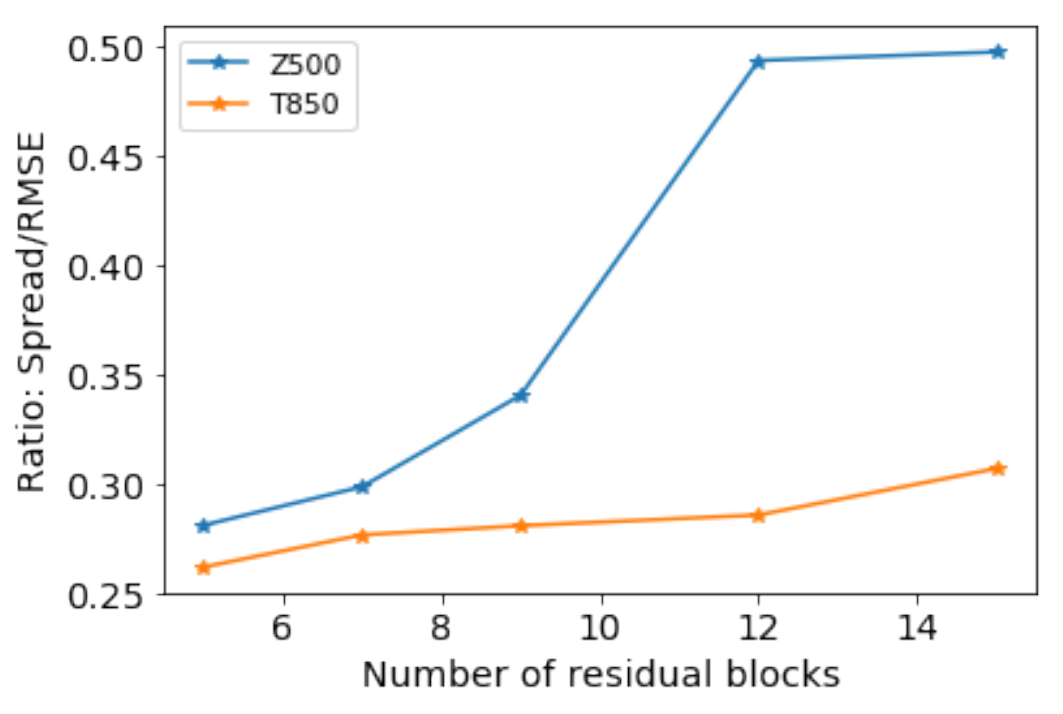}
		\caption{Ratio between the ensemble spread and the RMSE for the Z500 and T850 hindcasts. Note for a `perfect' ensemble, the ratio would be equal to 1.}\label{ensemble_spread}
	\end{figure}

	We also calculate the ensemble spread of the ensemble of Z500 and T850 outputs created using the dropout-at-inference technique. This is calculated using
	\begin{equation}
	\text{ensemble spread} = \sqrt{\mathbb{V}(y_{\text{ens}})},
	\end{equation}
	and is a measure of the uncertainty of the ensemble: if the ensemble is `perfect' then the ensemble spread should be equal to the RMSE (see \cite{palmer2006ensemble}). Figure \ref{ensemble_spread} shows the ratio between the ensemble spread for Z500 and T850 for the 3-day hindcast. Note in order to compare the spread and the RMSE directly, we have used the same weighted average from (\ref{RMSE}) on the spread to change it into a single value. Whilst our ensemble is not perfect, the ratio between the spread and the error is similar to that shown in Figure 3 in \cite{Scher2020}, when they use the dropout-at-inference technique to create an ensemble forecast of Z500. This shows that the use of the dropout ensemble technique here is appropriate and concludes our outline of the neural network architecture used in this work, which is summarised in schematic form in Figure \ref{resnet}.
	
	\subsubsection{Neural Network Stacking} \label{subsubsec:stacking}
	
	\begin{figure}[ht]
		\centering
		\includegraphics[width = 0.7\textwidth]{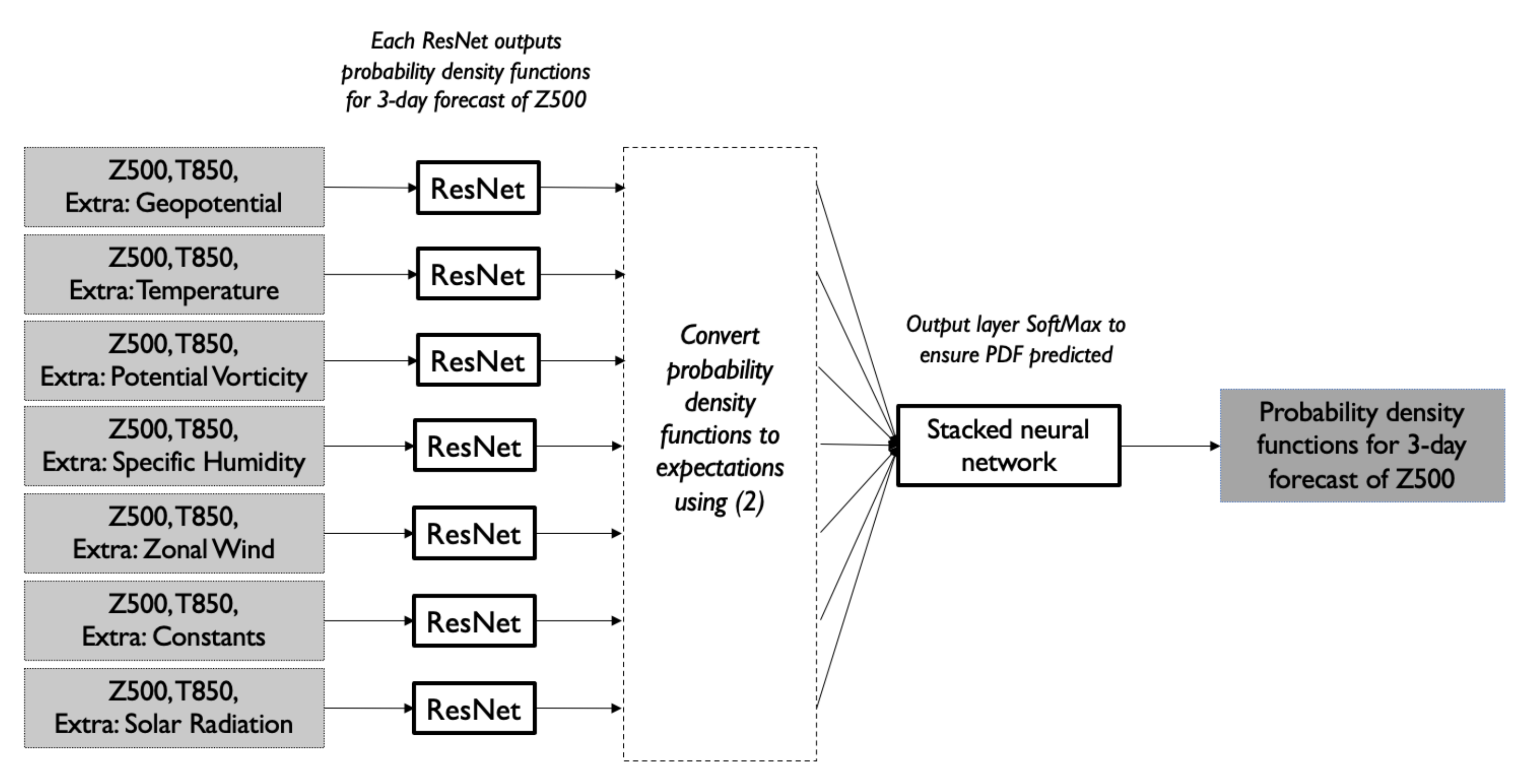}
		\caption[Schematic of stacked neural network]{Schematic for Z500 3-day hindcast using the stacked neural network approach to combine  outputs from individual ResNets (see Figure \ref{resnet} for their schematic). The schematics for the Z500 5-day, T850 3-day and T850 5-day hindcasts are the same except that there is no solar radiation ResNet for the T850 hindcasts. Note, each ResNet used in this set-up takes approximately 12 hours to train on a RTX6000 machine with two GPUs and 48GB of memory and the stacked neural network takes 30 minutes on the same machine.}
		\label{fig:schematic}
	\end{figure}
	
	Ideally, we would now train the neural network outlined above and shown in Figure \ref{resnet} on the WeatherBench dataset and predict Z500 and T850. However, the WeatherBench dataset is over 300GB and thus it is impractical and potentially unnecessary to train the neural network on the entire dataset at any one time. There are several methods to deal with this issue and in this work we focus on two: (i) using a meta-learner to combine the outputs from several neural networks trained individually; (ii) using data exploration techniques to identify the important variables in the dataset and discard the unimportant ones. In this section, we focus on constructing the meta-learner and then in Section \ref{subsec:data_exploration} we focus on the data exploration.
	
	In order to use a meta-learner, we set-up a series of ResNets (with the categorical data architecture in Figure \ref{resnet}) trained on smaller input datasets which always include Z500 and T850 as well as one other variable, all at current time $t$. This set-up is summarised on the left hand side of the schematic in Figure \ref{fig:schematic} and shows the other variables which make up the input datasets (note these variables are chosen following extensive analysis later in Section \ref{subsec:data_exploration}). The output of each of these networks is a density function at each point in space and time. To improve accuracy, for each network we use dropout at inference to create an ensemble of 32 outputs and use the law of total probability (\ref{eq:tot_prob}) to average the outputs meaning that the output of each individual neural network run is a single density function. Note, each individual ResNet takes approximately 12 hours to train on a RTX6000 machine with two GPUs and 48GB of memory, substantially less computational time and memory than would be required if only one neural network were used. 
	
	The outputs of these individual networks can now be combined to a single output for each point in time and space. There are several different methods to do this including linear pooling with average weights as done with the ensemble created by the dropout. Whilst linear pooling with average weights is sufficient for the dropout ensemble (because there is no reason why one ensemble member should be weighted higher than another), it is reasonable to assume some input variables are more important than others in determining the final results and thus should have a greater weighting (see Figures \ref{stacked_z500} and \ref{stacked_t850} in the next section). There are various specific techniques to combine outputs such as as pooling, voting and stacking \cite[]{combining_chapter}, as well as other simpler techniques such as linear regression. We choose to combine our outputs by using the learning technique of stacking \cite[]{wolpert1992stacked,smyth1999linearly} where the meta-learner is a stacked neural network used to combine individual learners (our individual ResNets). We make this choice because it is simple to implement and computationally cheap: the stacked neural network used in this work takes only 30 minutes to train on a RTX6000 machine with two GPUs and 48GB of memory. Moreover, combining distributions using techniques such as linear regression will almost definitely result in weightings which do not satisfy the law of total probability (\ref{eq:tot_prob}) and thus the combined output will not be a distribution. The advantage of using a stacked neural network approach is that by using a SoftMax layer as our output layer, we ensure that the combined output predicted by the stacked neural network is mathematically a distribution. This means that the inputs to our stacked neural network do not need to be distributions and thus to reduce memory and computational cost we transform the density functions from the individual neural networks to expectations using Eqn. (\ref{expectation}) before inputting them into the stacked neural network. This reduces the size of the input data into the stacked neural network by a factor of 100 due to there being 100 bins. Figure \ref{fig:schematic} provides a summary showing how the outputs from individual ResNets are combined using a stacked neural network. For the stacked neural network, we use a simple shallow network with the following architecture: an input layer to concatenate the output of the individual neural networks, two hidden layers with 36 nodes each and ReLU activations. A SoftMax output layer is used to generate a probability density function (as discussed above).
	
	\subsection{Data Exploration and Feature Selection}\label{subsec:data_exploration}
	In the previous section, we outline the full neural network architecture used in this work (summarised in Figures \ref{resnet} and \ref{fig:schematic}). However, a key component in the application of neural networks, or indeed any deep learning algorithm, is choosing appropriate data on which to train the model. This problem is two-fold: i) how many input variables and ii) how many individual data points to include. There is always a trade-off between having a large amount of data resulting in large computational and memory costs, and having too little data resulting in overfitting. Thus another method to improve the computational efficiency of our neural network is to perform data exploration to identify the important predictors for the forecasts.
	
	Before conducting formal data exploration, as a first step, we make some informed choices to only include certain variables in the potential input. Recall that the WeatherBench dataset contains a mix of multi-level variables and single level variables. We choose to include the multi-level variables of geopotential and temperature as these are the variables we are forecasting, and zonal wind because of its links to geopotential. We also choose to include a humidity multi-levelled variable and a vorticity multi-levelled variable. The WeatherBench dataset includes both specific humidity and relative humidity, and both potential vorticity and relative vorticity. From a physical perspective, we would expect little gain from using relative humidity as this variable is a function of specific humidity, temperature and pressure (which is itself related to geopotential), which are already present in the potential input dataset. Thus we choose specific humidity. Similarly, we exclude relative vorticity and keep potential vorticity because relative vorticity only describes the rotational component of the horizontal flow, whereas potential vorticity also includes a contribution from the vertical stratification of the temperature field and is known to be a conserved variable in adiabatic flow. In addition, potential vorticity is often used in the study of the development of mid-latitude weather systems and as a result is suitable for our potential input dataset. Finally, we also include the single level variables of solar radiation because it is the energy source driving the system; 2m temperature because of its likely influence on T850; and the constant fields of orography, land sea mask, and latitude.
	
	Thus, as a result of this initial physically-informed analysis, we have been able to exclude 45 variables from the training dataset out of a total of 115 (where for data on multiple levels, we are counting each individual level as a variable). For the remaining variables, we require further analysis.
	
	\subsubsection{Variable and level importance}\label{subsubsec:var_importance}
	After the physically-informed analysis, the set of relevant input variables consists of the multi-level fields of geopotential, temperature, specific humidity, potential vorticity and  zonal wind ($x$-direction) and the single-level fields of 2m temperature, top-of-atmosphere incoming solar radiation and constants (the three variables constant in time -- orography, land sea mask, and latitude). This gives a total of 70 single-level variables.  In this section, we conduct data exploration using a purely data-driven approach to further refine the set of relevant variables.
	
	In order to do so, we first train an individual neural network with an input dataset of just Z500 and T850, and define this as our \textquoteleft benchmark\textquoteright\hspace{0pt} training dataset, which will be used to understand the effect of including other variables on the final error. The architecture of this network is a ResNet (see Figure \ref{resnet}) with 5 residual blocks. We use the dropout at inference technique, described in Section \ref{subsubsec:dropout}, to extract an ensemble of 32 outputs from the single trained model, and then calculate the error of the ensemble mean (see Section \ref{subsubsec:dropout}). By using this, we average out some of the randomness of the results. 
	We then train individual neural networks with an input dataset of Z500 and T850 and one variable from the relevant list, and compare it to the original benchmark. Note that throughout this section, we have chosen to focus on improving the 3-day hindcast and assume that any improvements in methodology for this will also result in improvements in the 5-day hindcast. Thus unless explicitly stated all results shown are for the 3-day hindcast.
	
\begin{figure}[ht]
		\centering
		\includegraphics[width=0.4\textwidth]{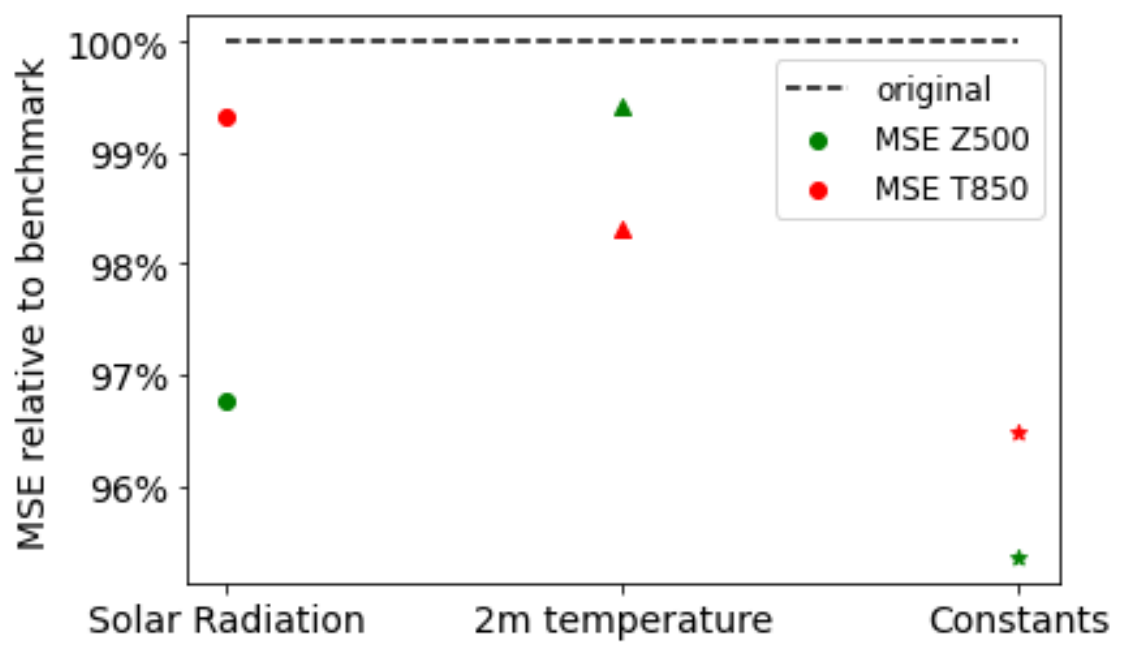}
		\caption{Percentage error relative to benchmark from training using extra single-level variables. These are: just incoming top-of-atmosphere solar radiation, just 2m temperature, or the three constants of orography, land-sea mask and latitude. Results are calculated from the mean of 32 ensemble members generated using dropout at inference, predicting on all gridpoints and times in the validation dataset (\textit{i.e.} 2016).}\label{fig:multidata}
	\end{figure}

\begin{figure}
		\begin{subfigure}{0.49\textwidth}
			\centering
			\includegraphics[width = 0.8\textwidth, height = 0.46\textwidth]{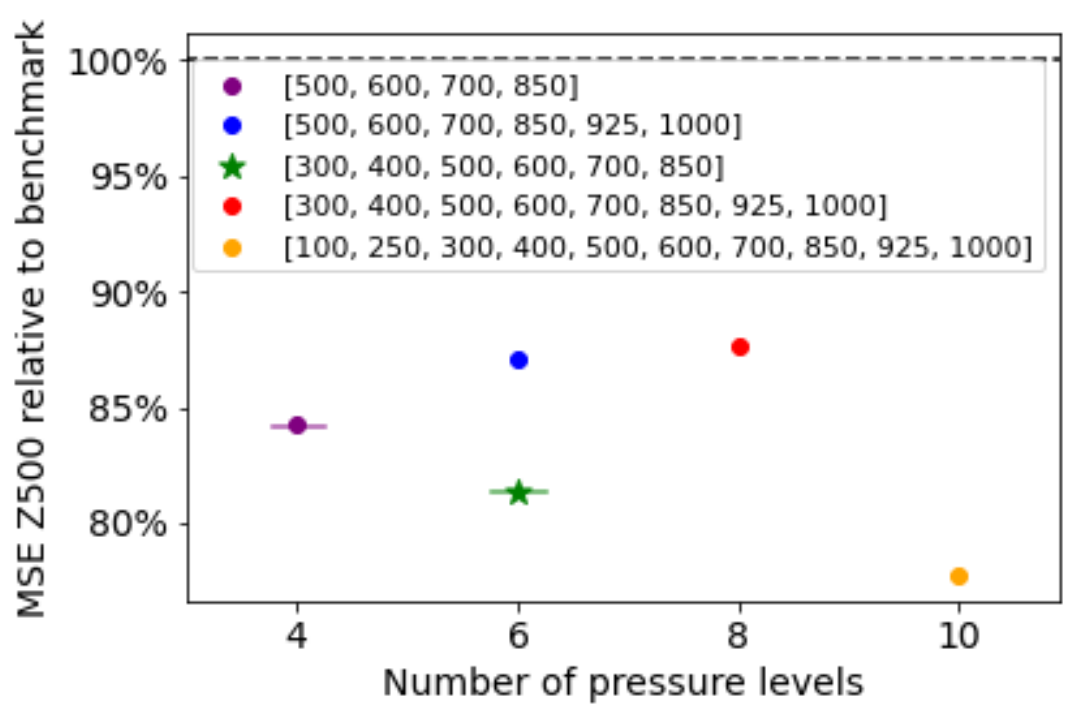}
			\caption{Geopotential: Error relative to benchmark for Z500.}
			\label{fig:z500_geo}
		\end{subfigure}
		\begin{subfigure}{0.49\textwidth}
			\centering
			\includegraphics[width = 0.8\textwidth, height = 0.46\textwidth]{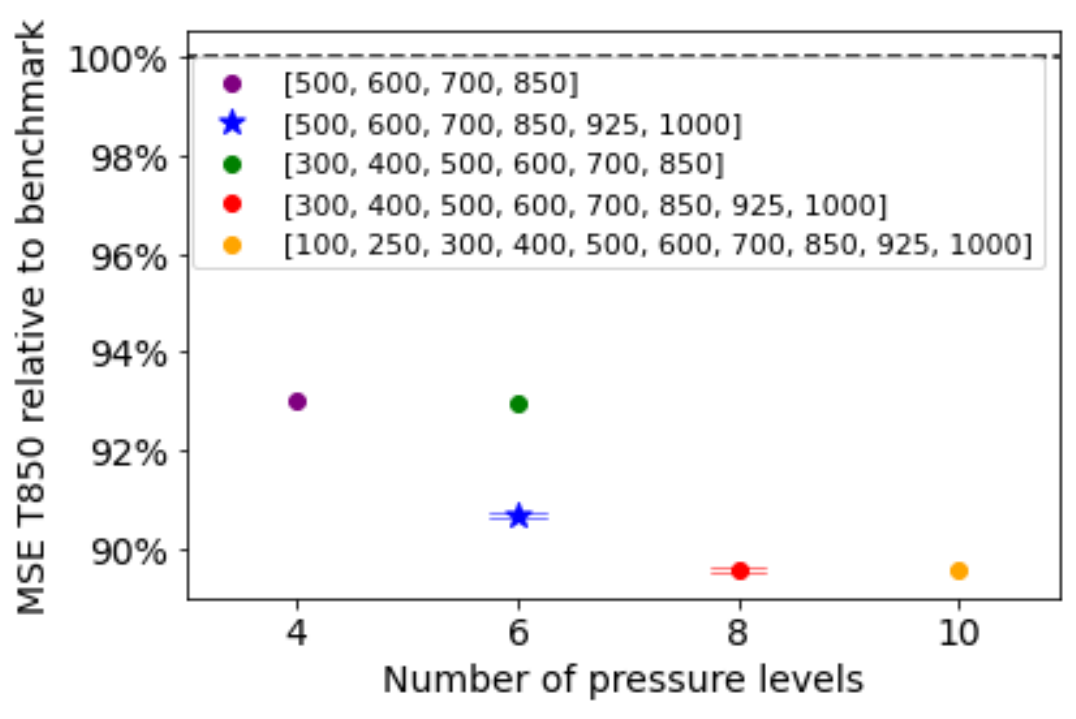}
			\caption{Geopotential: Error relative to benchmark for T850.}
			\label{fig:t850_geo}
		\end{subfigure}
		\begin{subfigure}{0.49\textwidth}
			\centering
			\includegraphics[width = 0.8\textwidth, height = 0.46\textwidth]{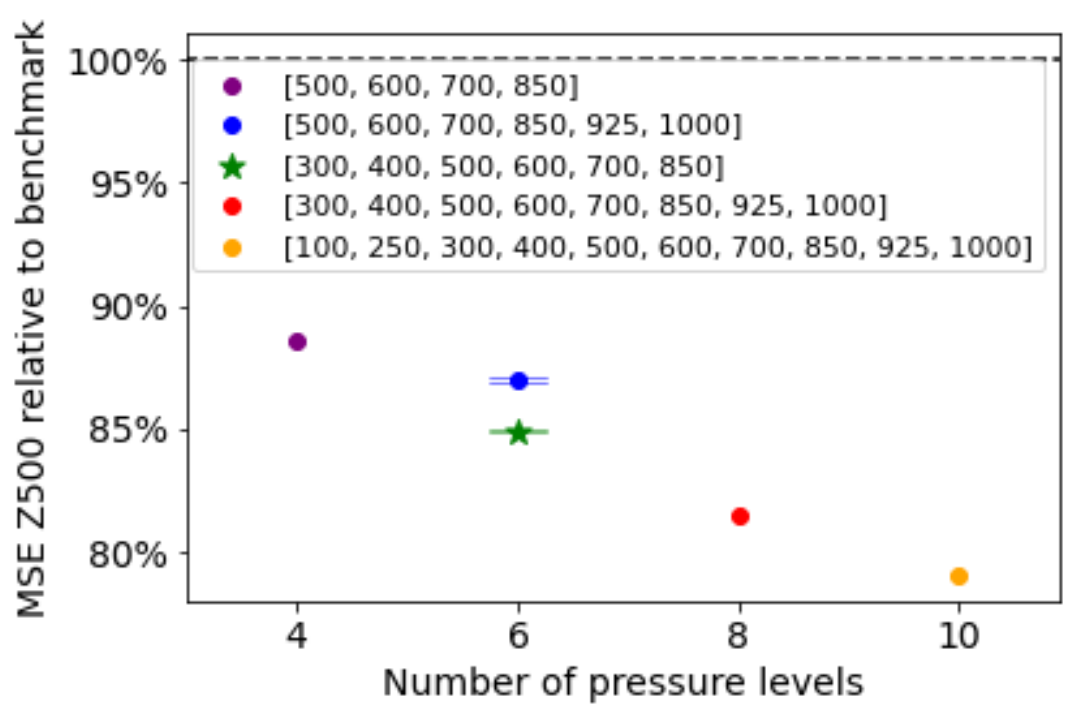}
			\caption{Temperature: Error relative to benchmark for Z500.}
			\label{fig:z500_temp}
		\end{subfigure}
		\begin{subfigure}{0.49\textwidth}
			\centering
			\includegraphics[width = 0.8\textwidth, height = 0.46\textwidth]{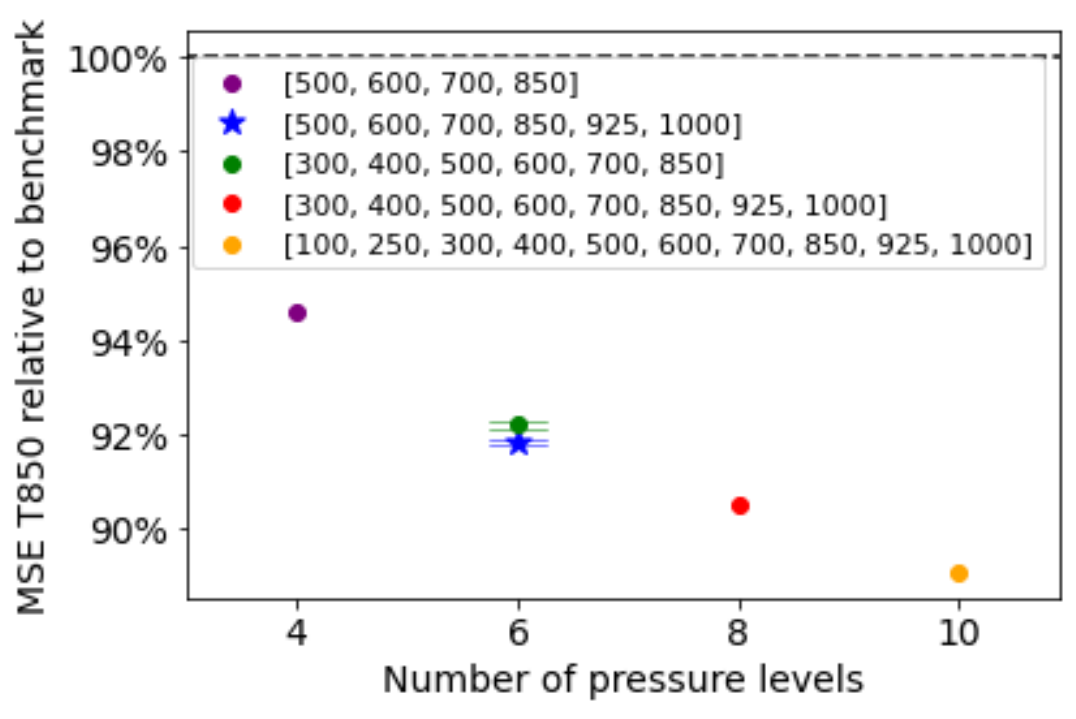}
			\caption{Temperature: Error relative to benchmark for T850.}
			\label{fig:t850_temp}
		\end{subfigure}
		\begin{subfigure}{0.49\textwidth}
			\centering
			\includegraphics[width = 0.8\textwidth, height = 0.46\textwidth]{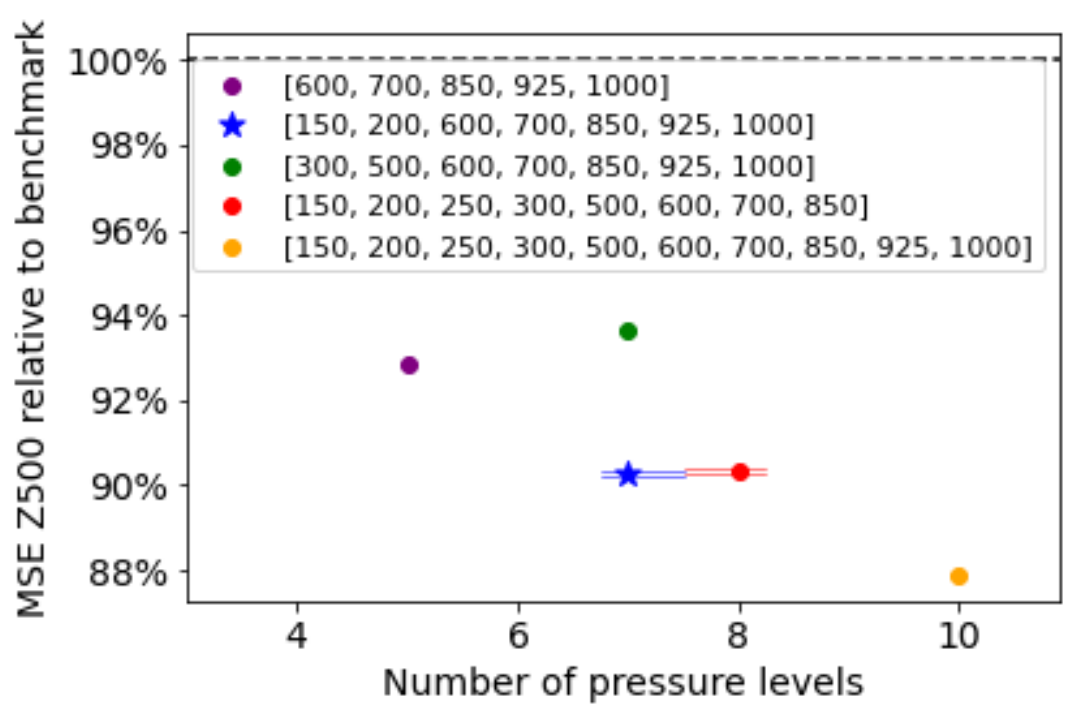}
			\caption{Specific humidity: Error relative to benchmark for Z500.}
			\label{fig:z500_sh}
		\end{subfigure}
		\begin{subfigure}{0.49\textwidth}
			\centering
			\includegraphics[width = 0.8\textwidth, height = 0.46\textwidth]{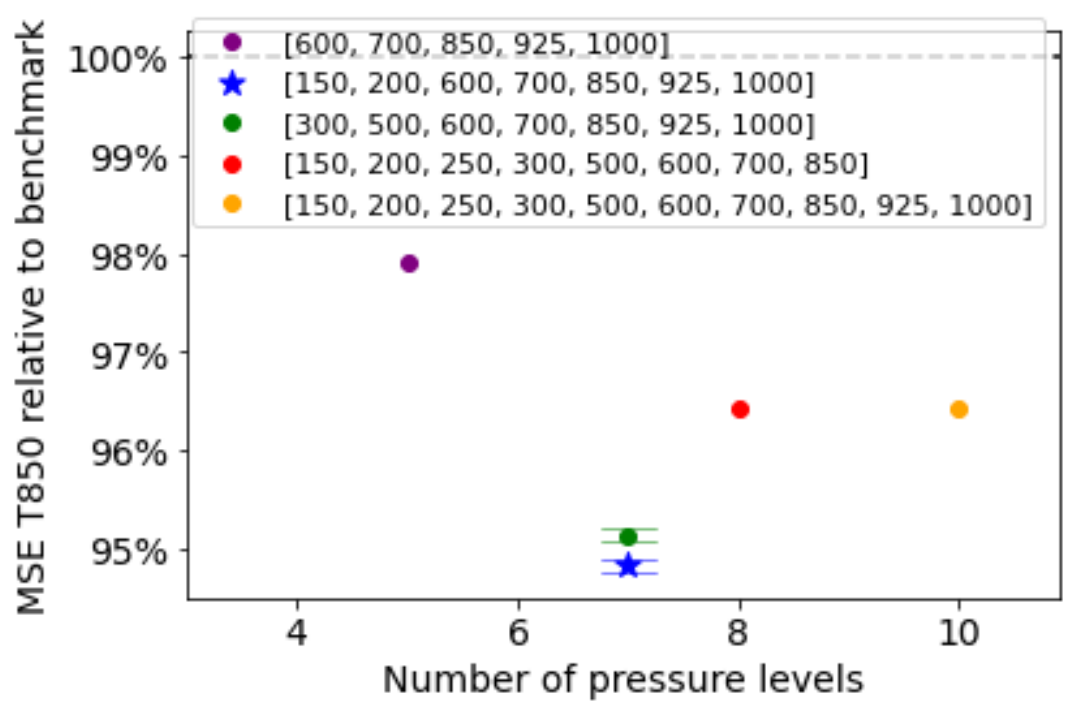}
			\caption{Specific humidity: Error relative to benchmark for T850.}
			\label{fig:t850_sh}
		\end{subfigure}
		\begin{subfigure}{0.49\textwidth}
			\centering
			\includegraphics[width = 0.8\textwidth, height = 0.46\textwidth]{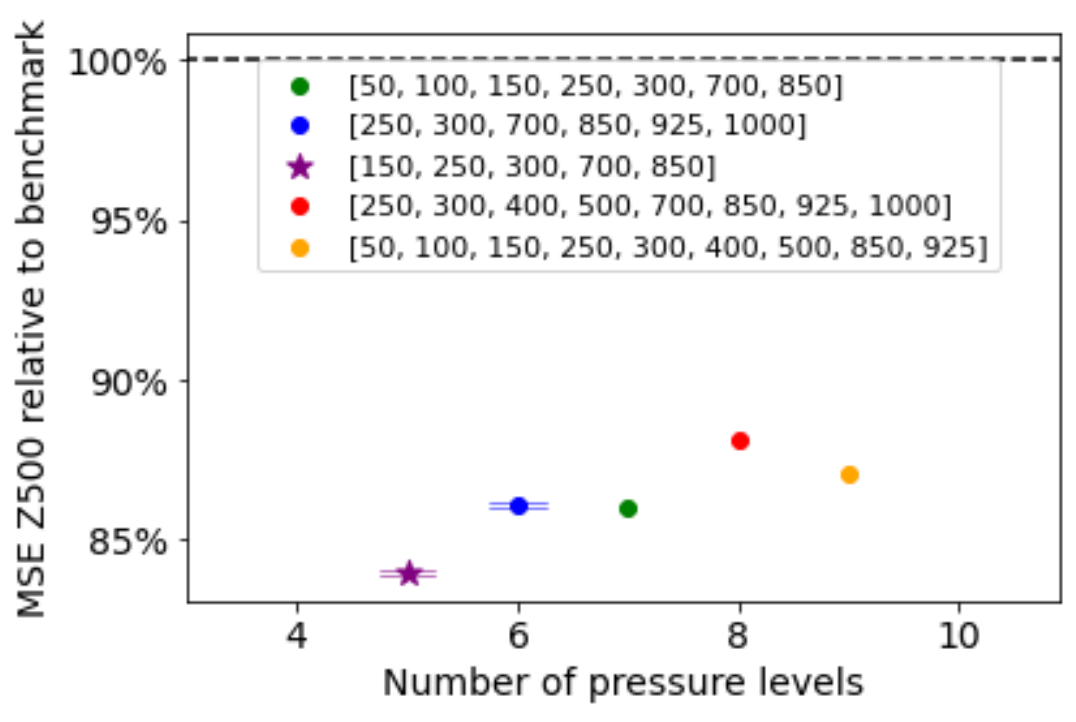}
			\caption{Potential vorticity: Error relative to benchmark for Z500.}
			\label{fig:z500_pv}
		\end{subfigure}
		\begin{subfigure}{0.49\textwidth}
			\centering
			\includegraphics[width = 0.8\textwidth, height = 0.46\textwidth]{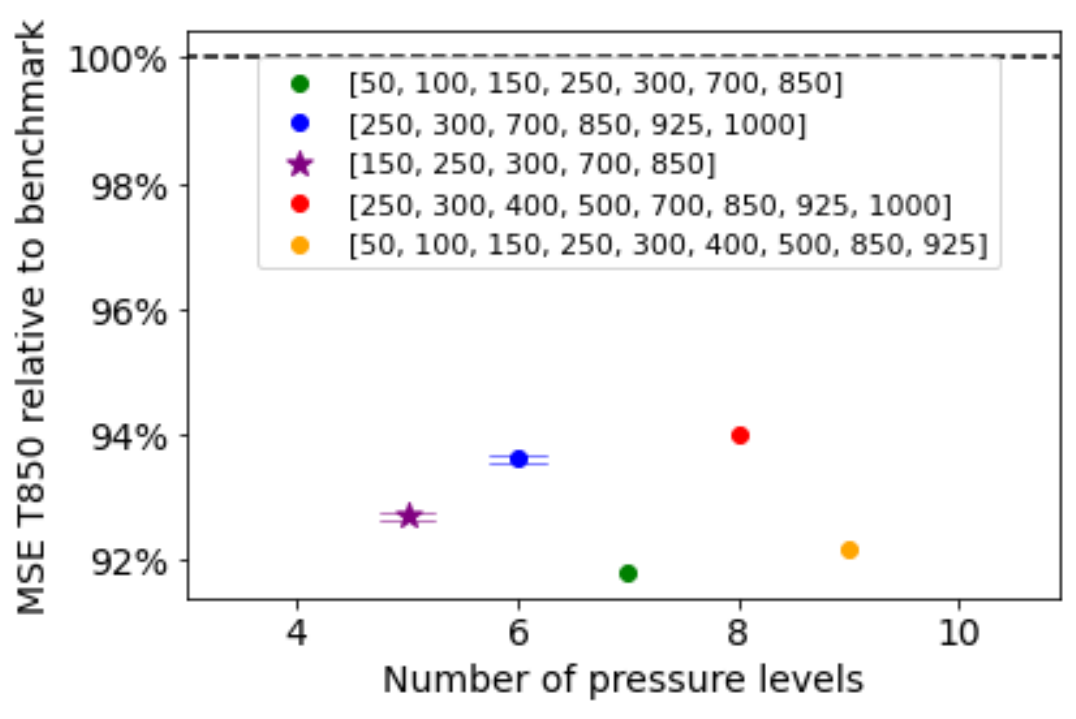}
			\caption{Potential vorticity: Error relative to benchmark for T850.}
			\label{fig:t850_pv}
		\end{subfigure}
		\begin{subfigure}{0.49\textwidth}
			\centering
			\includegraphics[width = 0.8\textwidth, height = 0.46\textwidth]{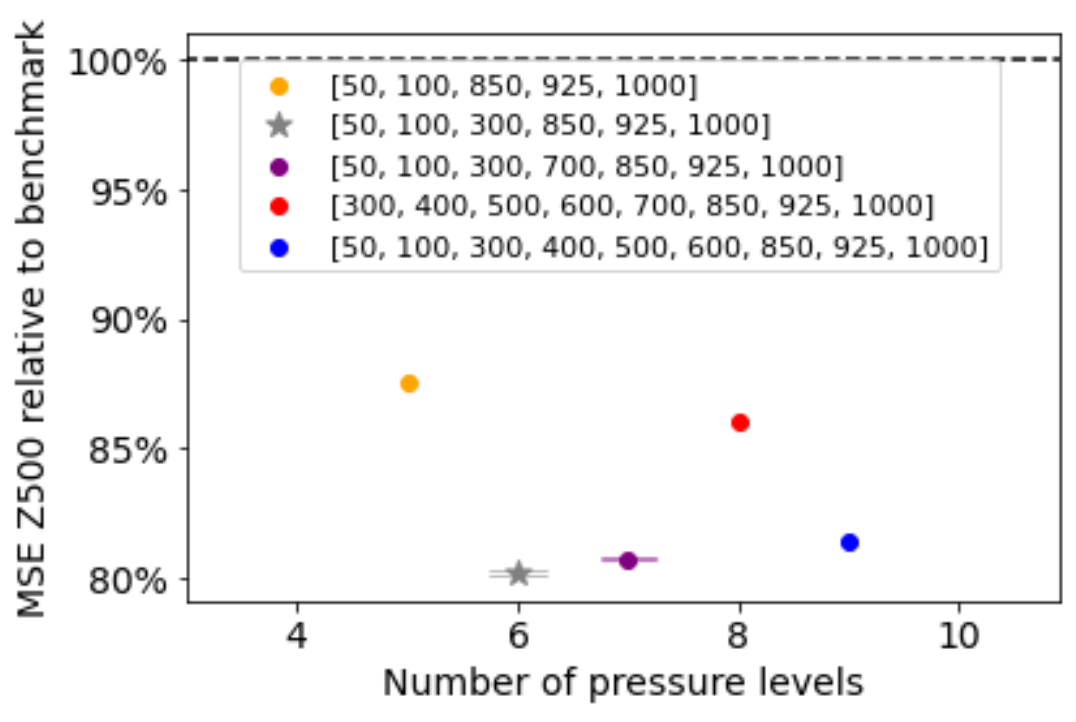}
			\caption{Zonal wind: Error relative to benchmark for Z500.}
			\label{fig:z500_wind}
		\end{subfigure}
		\begin{subfigure}{0.49\textwidth}
			\centering
			\includegraphics[width = 0.8\textwidth, height = 0.46\textwidth]{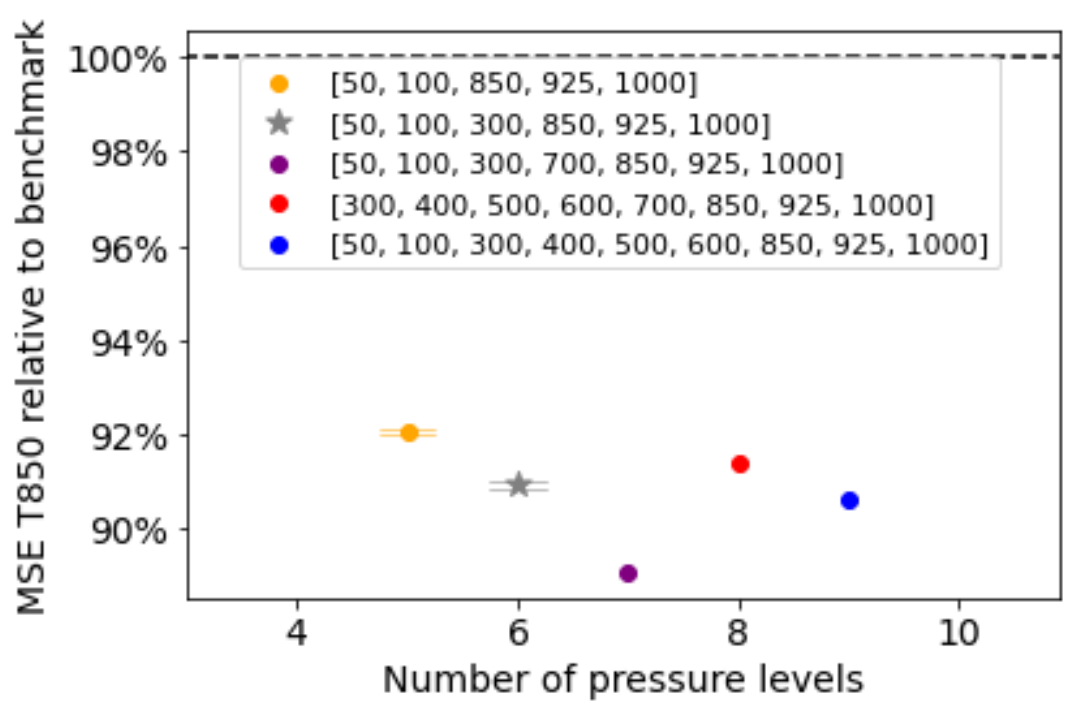}
			\caption{Zonal wind: Error relative to benchmark for T850.}
			\label{fig:t850_wind}
		\end{subfigure}
		\caption{Percentage error relative to benchmark from training using extra variable levels (chosen levels are denoted by a star). The 95\% confidence interval bars are shown for some level combinations. Some intervals are so narrow that it is not possible to distinguish between the upper and lower error bars on the figure, but we emphasise here that all error bars are in the $y$-direction. Results are the mean of 32 ensemble members, predicting on all gridpoints and times in the validation dataset (\textit{i.e.} 2016).} \label{fig:levels}
	\end{figure}

	We start by considering the significance of the variables which are only defined at a single height: 2m temperature, total incident solar radiation and constants. The resultant ensemble errors are shown in Figure \ref{fig:multidata} as a percentage of the error relative to the benchmark dataset. The figure shows that solar radiation is relatively unimportant for predicting T850 and relatively important for predicting Z500, that 2m temperature is relatively unimportant for predicting Z500 and T850 and finally constants are important for predicting both Z500 and T850. Thus from Figure \ref{fig:multidata}, we conclude that we can exclude 2m temperature from both training datasets and solar radiation from the T850 training dataset. Note all relative errors in this figure are relative mean squared errors (MSE) so that these results can be compared with those in Figure \ref{fig:levels} which uses MSE to make calculating confidence intervals easier (see later in this section).
	
	We next seek to understand which pressure levels are the most important in predicting Z500 and T850 using a purely data-driven approach, which often agrees with well-established theory in meteorology, as noted in Section \ref{sec:physical}. For all multi-level input variables (geopotential, temperature, specific humidity, potential vorticity and zonal wind), we consider multiple different level combinations. We use these combinations to train a 5 block convolutional ResNet and, as before, use the dropout at inference technique (described in Section \ref{subsubsec:dropout}) to extract an ensemble of 32 outputs from the single trained model, and then calculate the error of the ensemble mean (see Section \ref{subsubsec:dropout}).
	
	The statistical significance of the multi-level analysis  results is analysed by calculating the 95\% confidence interval of the MSE of the hindcast at each point in space and time in the validation dataset. We use the MSE rather than the RMSE for ease of calculation, and therefore the 95\% confidence interval formula is
	\begin{equation}\label{95_conf}
	\text{MSE} \pm  1.96 \hspace{2pt}\sqrt{\frac{\mathbb{V}\left[L(j)(f_{i,j,k}-t_{i,j,k})^{2}\right]}{N}},
	\end{equation}
	where 
	\begin{equation}
	\text{MSE} =\mathbb{E}\left[L(j)(f_{i,j,k}-t_{i,j,k})^{2}\right] = \frac{1}{N_{\text{lat}}N_{\text{lon}}N_{\text{timepoints}}} \sum_{i=1}^{N_{\text{timepoints}}}\sum_{j=1}^{N_\text{lat}}\sum_{k=1}^{N_\text{lon}}L(j)(f_{i,j,k}-t_{i,j,k})^{2},
	\end{equation}
	$L(j)$ is the latitude weighting factor given by (\ref{Lij}), $f$ is the predicted value from the neural network, $t$ is the true value from the dataset and $N$ is the total number of points in space and time in the validation dataset (equal to $N_{\text{timepoints}} \times N_{\text{lat}} \times N_{\text{lon}}$). 
	
	The results of the level analysis are shown in Figure \ref{fig:levels} as percentages relative to the benchmark. For brevity and figure clarity, the confidence intervals are shown only for the optimum level choice (denoted by a star) and to another level combination with an error close to that of the optimum error choice. As the confidence intervals are so narrow, the error bars are extended to make them clearer. Even so, in some cases, the confidence intervals are so narrow that it is not possible to distinguish between the upper and lower error bars on the figure, making it difficult to interpret them. To avoid confusion, we emphasise here that all error bars shown in these figures are in the $y$-direction. The remainder of this section provides more detail on how the optimum level choice for each variable is determined.
	
	The optimum levels for the geopotential and temperature variables are determined in a systematic way. First we consider the two levels we are trying to predict and the levels between these two (\textit{i.e.} [500hPa, 600hPa, 700hPa, 850hPa]); next we extend this dataset by including two levels with larger pressure values (\textit{i.e.} [500hPa, 600hPa, 700hPa, 850hPa, 925hPa, 1000hPa]);
	then we add the two levels with smaller pressure values to the first dataset  (\textit{i.e.} [300hPa, 400hPa, 500hPa, 600hPa, 700hPa, 850hPa]); then we include both the levels with larger and smaller pressure values (\textit{i.e.} [300hPa, 400hPa, 500hPa, 600hPa, 700hPa, 850hPa, 950hPa and 1000hPa]); and finally we add two further levels with small pressure values (\textit{i.e.} [100hPa, 250hPa, 300hPa, 400hPa, 500hPa, 600hPa, 700hPa, 850hPa, 950hPa and 1000hPa]). The results of this analysis are shown in Figures \ref{fig:z500_geo} and \ref{fig:t850_geo} for the geopotential variable inputs and  \ref{fig:z500_temp} and \ref{fig:t850_temp} for the temperature variable inputs. In general, they show for both Z500 and T850 that including more pressure levels results in a larger decrease in error relative to the benchmark, but this must be balanced with computational cost. When predicting Z500, a good compromise for both the geopotential and temperature input variables is the level choice [300hPa, 400hPa, 500hPa, 600hPa, 700hPa, 850hPa] (green stars on Figures \ref{fig:z500_geo} and \ref{fig:z500_temp}). Figures \ref{fig:z500_geo} and \ref{fig:z500_temp} show the 95\% confidence interval for this optimum level choice does not overlap with that of the level choice with the most similar accuracy, and the same result applies for all other level choices (not shown for brevity). Therefore, our optimum level choice is statistically significantly different in accuracy to all other level choices considered.  When predicting T850, a good compromise for both the geopotential and temperature input variables is the level choice [500hPa, 600hPa, 700hPa, 850hPa, 925hPa, 1000hPa] (blue stars on Figures \ref{fig:t850_geo} and \ref{fig:t850_temp}). Like when predicting Z500, Figures \ref{fig:t850_geo} and \ref{fig:t850_temp} show no overlapping of the 95\% confidence interval (\ref{95_conf}), and therefore our optimum level choice is statistically significantly different in accuracy. Given that T850 has a larger pressure value than Z500, it is unsurprising that the T850 level choice contains levels with larger pressure values than the Z500 level choice.
	
	For the other variables, the same analysis does not apply and we conducted a small correlation analysis between the different levels of specific humidity, potential vorticity and zonal wind, and the target variables using pattern correlation to determine which levels might be good predictors. This analysis is not included here for brevity but is available in the data exploration section of the GitHub repository (see Section \ref{code}). For specific humidity, this analysis showed that the correlation is greatest at the levels with large pressure values. Thus, we begin by considering the levels with larger pressure values (\textit{i.e.} [600hPa, 700hPa, 850hPa, 925hPa, 1000hPa]), then add different subsets of levels with small pressure values, and finally consider a broad range of pressure levels (\textit{i.e.} [150hPa, 200hPa, 250hPa, 300hPa, 500hPa, 600hPa, 700hPa, 850hPa, 925hPa, 1000hPa]). Figures \ref{fig:z500_sh} and \ref{fig:t850_sh} show the results of this analysis and that, unlike with the temperature and geopotential input variables, adding more pressure levels does not always result in the error decreasing relative to the benchmark. Without clear correlation between the number of pressure levels and the error reduction, we make the level choice based on reducing the error while keeping the number of levels to as few as possible and thus choose [150hPa, 200hPa, 600hPa, 700hPa, 850hPa, 925hPa, 1000hPa] (blue stars on Figures \ref{fig:z500_sh} and \ref{fig:t850_sh}) for both Z500 and T850 prediction. Figure \ref{fig:t850_sh} shows that when predicting T850, the 95\% confidence interval (\ref{95_conf}) of this optimum level choice does not overlap with that of the level choice with the most similar accuracy, showing that our optimum level choice is statistically significantly different in accuracy. However, Figure \ref{fig:z500_sh} shows that when predicting Z500, the difference in the accuracy achieved by using [150hPa, 200hPa, 600hPa, 700hPa, 850hPa, 925hPa, 1000hPa] (our optimum level choice) compared to [150hPa, 200hPa, 250hPa, 300hPa, 500hPa, 600hPa, 700hPa, 850hPa] is not statistically significant (95\% confidence intervals overlap). This means we cannot use the criteria of accuracy to distinguish between the two level combinations, but the difference in computational cost between the two level combinations still justifies our optimum level choice.
	
	For potential vorticity, the correlation analysis showed that the most important levels are at either end of the pressure level spectrum. Thus when determining the right level choice, we consider groupings of levels at both large and small pressure values. We also consider the middle pressure levels shifted to the larger end of the pressure spectrum (\textit{i.e.} [250hPa, 300hPa, 400hPa, 500hPa, 700hPa, 850hPa, 925hPa, 1000hPa]) and to the smaller end of the pressure spectrum (\textit{i.e.} [50hPa, 100hPa, 150hPa, 250hPa, 300hPa, 400hPa, 500hPa, 850hPa, 925hPa]). Figures \ref{fig:z500_pv} and \ref{fig:t850_pv} show that there does not seem to be a clear pattern between level choice and error decrease for potential vorticity. However, the level choice of [150hPa, 250hPa, 300hPa, 700hPa, 850hPa] (purple stars on Figures \ref{fig:z500_pv} and \ref{fig:t850_pv}) results in a large error decrease in both RMSE Z500 and RMSE T850 and is also relatively computationally cheap and so is an appropriate level choice for potential vorticity. Moreover, the figures show that, for both Z500 and T850, the 95\% confidence interval (\ref{95_conf}) of this optimum level choice does not overlap with the confidence interval of the level choice closest in accuracy to it, which also applies for all other level choices (not shown for brevity). This means our optimum level choice is statistically significantly different in accuracy to the other level choices.
	
	Finally we consider the zonal wind input variable. Using the correlation analysis, the most important levels are at either end of the pressure spectrum. Thus we consider groupings of levels at both large and small pressure values. We also consider groupings at the larger end of the pressure spectrum (\textit{i.e.} [300hPa, 400hPa, 500hPa, 600hPa, 700hPa, 850hPa, 925hPa, 1000hPa]) to check if the smaller pressure values are affecting the results. Figures \ref{fig:z500_wind} and \ref{fig:t850_wind} show that for both Z500 and T850 having levels at both small and large pressure values is important for reducing error. In both cases, we choose the level grouping of [50hPa, 100hPa, 300hPa, 850hPa, 925hPa, 1000hPa] (grey stars on Figures \ref{fig:z500_wind} and \ref{fig:t850_wind}) as an appropriate choice because it results in a large error reduction relative to the benchmark and is also relatively computationally cheap. Figure \ref{fig:z500_wind} shows that the 95\% confidence interval (\ref{95_conf}) of this optimum level choice does not overlap with the confidence interval of the level choice closest in accuracy to it, meaning our level choice is statistically significantly different in accuracy to the other level choice. In the case of Figure \ref{fig:t850_wind} we chose to compare the 95\% confidence interval of the optimum level choice with that of [50hPa, 100hPa, 850hPa, 925hPa, 1000hPa] because of the higher computational cost of the level choice which is closest in accuracy. The confidence intervals do not overlap, hence our optimum level choice is statistically significantly different in accuracy compared to the computationally cheaper option.
	
	Thus in this section, we have conducted a data-driven level importance analysis and shown that our optimum level choices are always statistically significantly different from other level choices of a lower or equivalent computational cost, and are almost always statistically significantly different from other more computationally expensive level choices.
	
	\subsubsection{Agreement with meteorological theory}\label{sec:physical}
	So far, the choice of which pressure levels to include has been done in a predominantly data-driven manner. It is worth emphasizing that the focus has been on predicting Z500 and T850 three days ahead. If other target variables, or other lead-times, had been the focus of the predictions, then other input levels may have been found to be beneficial. Having found the most useful input levels, it is now sensible to look back and comment on whether the choices  are supported by physical reasoning. This is the case for many of the pressure level choices that we have made:
	\begin{itemize}
		\item Geopotential \& Temperature:  Using physical intuition, we would expect that the levels that are most important for prediction are those above and below the level of interest \cite[e.g.][]{hoskins1978new}.
		\item Potential vorticity: Our training dataset contains levels both near the tropopause (e.g. 150 hPa) and just above the boundary-layer (e.g. 850 hPa). We would expect these to be important because they allow the neural network to learn about the interactions between lower and upper level potential vorticity which can accelerate cyclone development \citep[e.g.][]{hoskins}. 
		\item Zonal wind: Quasi-Geostrophic Theory \cite[]{pedder1997omega} tells us that in order to determine regions of rising and falling air, we require knowledge of the wind fields at high altitude (e.g. 50hPa) and low altitude (e.g. 1000hPa), meaning our choice of levels is physically reasonable. 
	\end{itemize}
	It should be noted, however, that just using physical reasoning alone would not have been sufficient to determine which pressure levels are good predictors. Firstly, in the case of specific humidity, we are unaware of any clear reasoning as to whether the important pressure levels should be in low or high pressure regions. Furthermore, for other variables, whilst physical reasoning may suggest whether low and/or high pressure regions are important, the WeatherBench dataset has a number of levels in `low' or `high' pressure regions. Our data driven approach allows us to determine which and how many of these levels to choose, whereas physical reasoning does not.
	
	This agreement with physical reasoning shows that neural network based feature selection is able to identify physically important inputs, which include highly correlated time series data. The different pressure levels of a single variable are particularly highly correlated with each other and the neural network based feature selection is able to identify which pressure levels are important for prediction. Both the agreement with key physical principles and the ability to deal with correlated data are identified in \cite{Schultz2021} as areas where neural networks need to prove themselves in order to be able to compete with numerical weather prediction models. Thus this good agreement with physical reasoning validates our neural network approach.
	
	In summary, in this section we have determined a set of input variables which are important in determining Z500 and T850. From an original dataset of 115 variables, we have been able to exclude 45 through an initial physically-informed analysis and a further 36 for Z500 and 37 for T850 through our variable and level importance analysis. We use this input variable selection in the next section to reduce computational and memory costs without compromising too much on accuracy.
	
	\section{Results}\label{sec:results}
	In the previous section, we outlined our new methodology for predicting the weather forecast using a data-driven approach. In this section, we show the results of implementing this methodology. Because we are now applying the full stacked neural network approach, we must split the dataset in a different way to in Section \ref{sec:methods} as there are now two stages of neural networks. We use the data from 1979 to 2011 as the training dataset for the individual ResNet learners, and use the data from 2011 as a validation dataset. We then use the individual learners to make predictions on the dataset from 2012 to 2016 (hereafter referred to as the stacked validation dataset). These predictions are the inputs for the meta-learner (the stacked neural network), which uses shuffle as validation. As in \cite{Rasp2020a}, the final testing dataset is the data from 2017 to 2018. As discussed previously, all the analysis so far has been carried out for the 3-day weather hindcast but we show that this methodology also leads to good results for the 5-day weather hindcast.
	
\begin{figure}[ht]
		\begin{subfigure}{0.49\textwidth}
			\centering
			\includegraphics[width =0.9\textwidth]{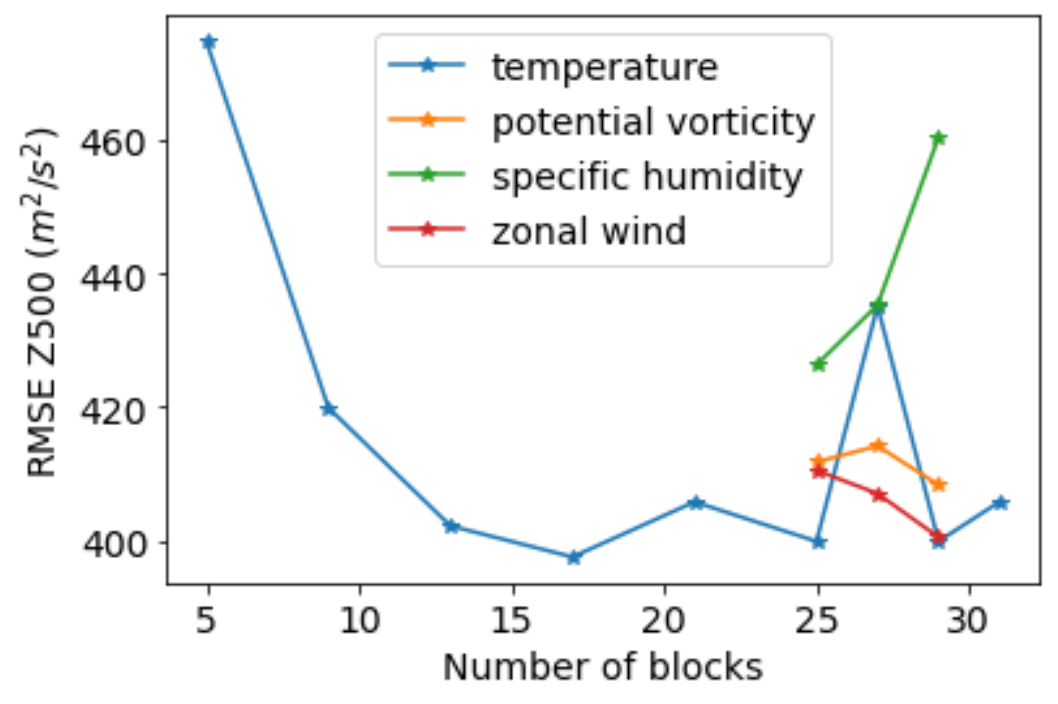}
			\caption{RMSE for 500hPa geopotential.}
			\label{fig:opt_res_z500}
		\end{subfigure}
		\begin{subfigure}{0.49\textwidth}
			\centering
			\includegraphics[width =0.9\textwidth]{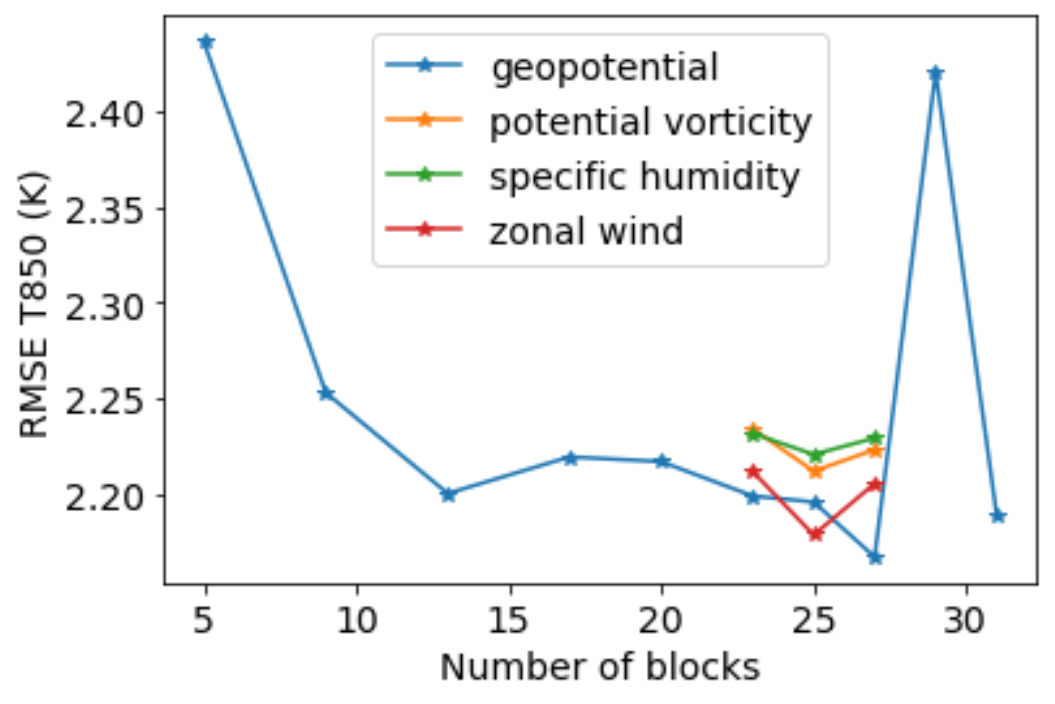}
			\caption{RMSE for 850hPa temperature.}
			\label{fig:opt_res_t850}
		\end{subfigure}
		\caption[RMSE comparison]{RMSE as a result of using different numbers of residual blocks in the ResNet. The effect on the error is shown for ResNets trained on a number of different variables. The RMSE is calculated using the weighted average (\ref{RMSE}) over all gridpoints and times in the stacked validation dataset (\textit{i.e.} 2012--2016).}
	\end{figure}

	As a first step, we determine the optimum number of residual blocks to use in the individual ResNets which feed the stacked neural network. (Recall that figures \ref{categorical} and \ref{comp_do} in the previous sections show that the accuracy of a ResNet is very dependent on the number of residual blocks used). Figure \ref{fig:opt_res_z500} and \ref{fig:opt_res_t850} show how RMSE Z500 and RMSE T850 vary as the number of residual blocks is increased by steps of four blocks from 5 to 25 blocks and steps of two blocks from 25 to 31. The error is calculated on the 2012 to 2016 dataset, because these are the predictions which are used as the input for the stacked neural network. For the main analysis of the optimum number of blocks, we use the optimum temperature levels as the training dataset for Z500 and the optimum geopotential levels as the training dataset for T850. Before 25 residual blocks, in both cases the error decreases relatively steadily as the number of residual blocks increases. However, beyond this number the error sometimes increases dramatically when more residual blocks are added and sometimes decreases dramatically. Thus we also conduct a small sensitivity analysis using the optimum levels of potential vorticity, specific humidity and zonal wind. Figure \ref{fig:opt_res_z500} shows the error is minimised with 29 residual blocks for temperature, potential vorticity and zonal wind, but for specific humidity the error increases notably when 29 blocks are used. Given that the error using specific humidity is always higher than that from using the other variables, the contribution from the specific humidity variable after the stacked neural network is applied will be small. Thus it is reasonable to ignore that the error from specific humidity is high when 29 blocks are used and conclude 29 is a good number of residual number of blocks to use for Z500. Figure \ref{fig:opt_res_t850} shows the error is minimised with 25 blocks for potential vorticity, specific humidity and zonal wind and close to being minimised by this number of blocks for geopotential. Thus, we conclude 25 is a good number of residual blocks to use for T850.
	
\begin{figure}[ht]
		\begin{subfigure}{0.49\textwidth}
			\centering
			\includegraphics[width =0.9\textwidth]{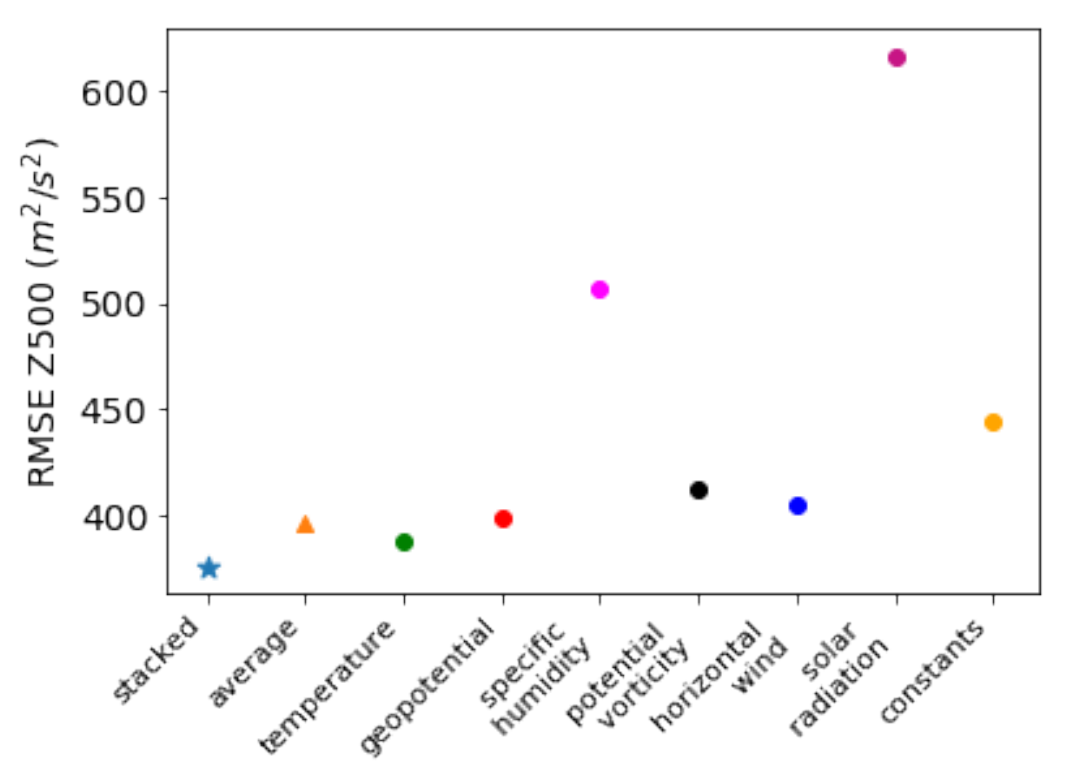}
			\caption{Error for 500hPa geopotential \\ \hspace{30pt}(ResNets with 29 residual blocks).}
			\label{fig:stacked_z500_3days}
		\end{subfigure}
		\begin{subfigure}{0.49\textwidth}
			\centering
			\includegraphics[width =0.9\textwidth]{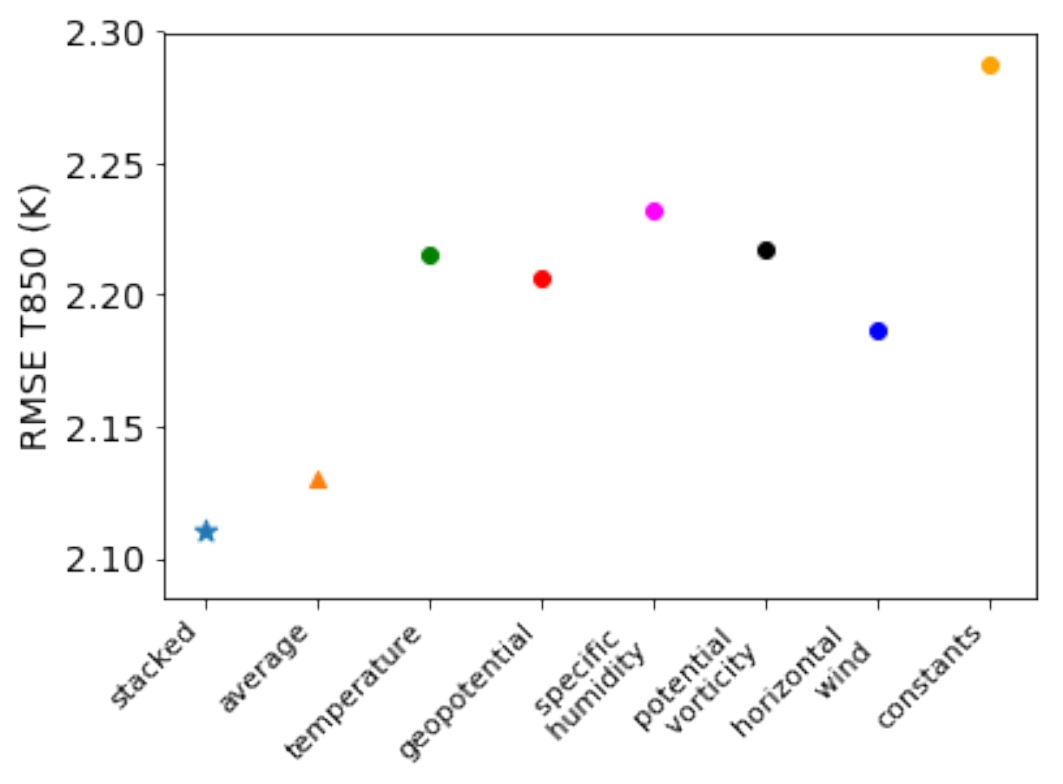}
			\caption{Error for 850hPa temperature \\ \hspace{30pt}(ResNets with 25 residual blocks).}
			\label{fig:stacked_t850_3days}
		\end{subfigure}
		\caption[Improvement]{Improvement in accuracy for the 3-day hindcast as a result of using a stacked neural network compared to the individual ResNets trained on specific variables and simply averaging the outputs of the individual ResNets. The RMSE is a weighted average calculated using (\ref{RMSE}) over all gridpoints and times in the test dataset (\textit{i.e.} 2017--2018).}\label{stacked_z500}
	\end{figure}

\begin{figure}[ht]
		\begin{subfigure}{0.49\textwidth}
			\centering
			\includegraphics[width =0.9\textwidth]{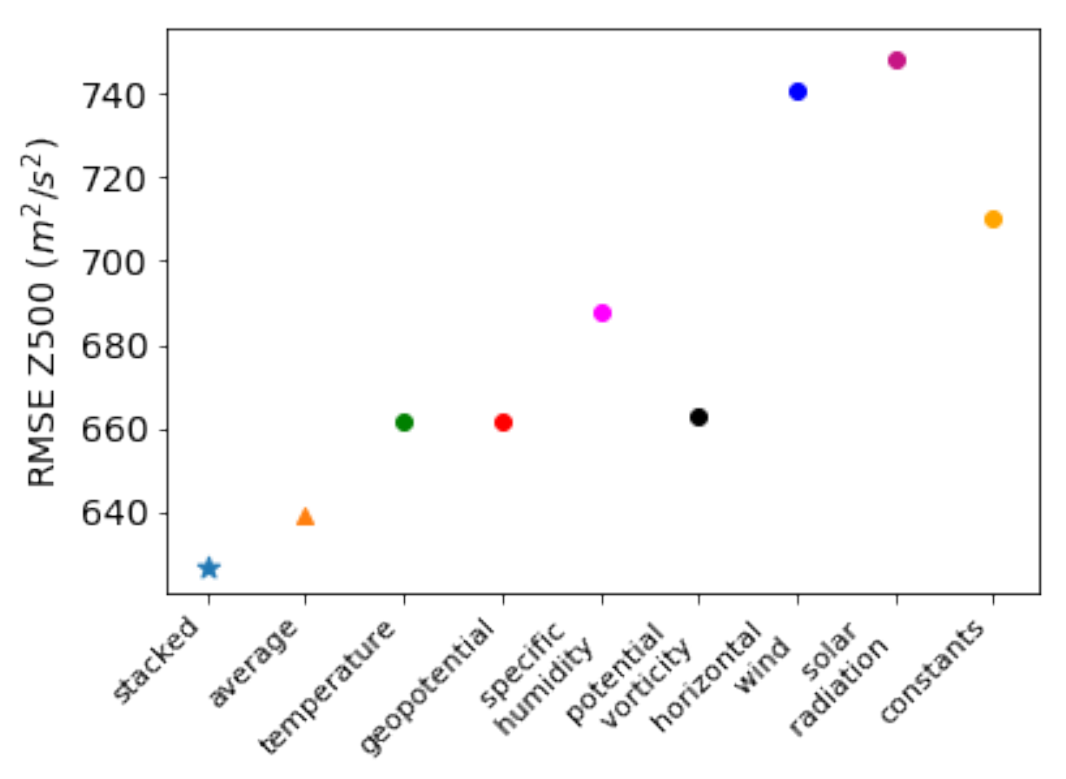}
			\caption{Error for 500hPa geopotential\\ \hspace{30pt}(ResNets with 29 residual blocks).}
			\label{fig:stacked_z500_5days}
		\end{subfigure}
		\hfill
		\begin{subfigure}{0.49\textwidth}
			\centering
			\includegraphics[width =0.9\textwidth]{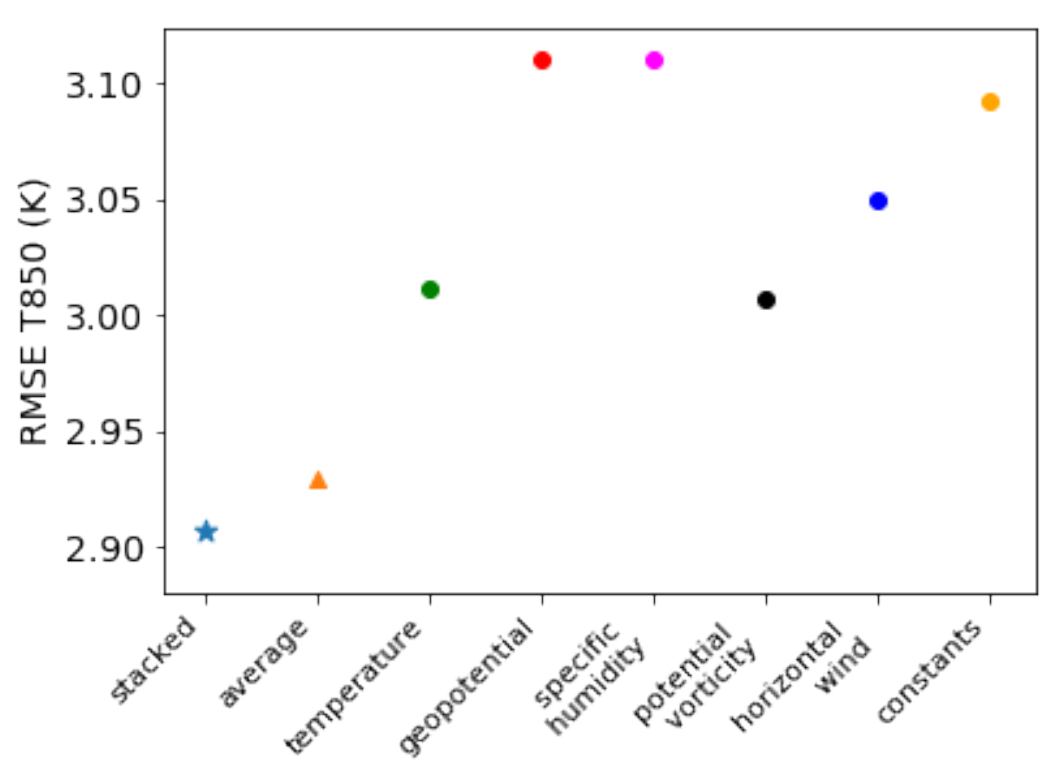}
			\caption{Error for 850hPa geopotential \\ \hspace{30pt}(ResNets with 25 residual blocks).}
			\label{fig:stacked_t850_5days}
		\end{subfigure}
		\caption[Improvement 5 day]{Improvement in accuracy for the 5-day hindcast as a result of using a stacked neural network compared to the individual ResNets trained on specific variables and simply averaging the outputs of the individual ResNets. The RMSE is a weighted average calculated using (\ref{RMSE}) over all gridpoints and times in the test dataset (\textit{i.e.} 2017--2018).}\label{stacked_t850}
	\end{figure}

	Using these optimum numbers of residual blocks, we can now calculate our full neural network results. Figures \ref{fig:stacked_z500_3days} and \ref{fig:stacked_t850_3days} compare, for Z500 and T850 respectively, the 3-day prediction error on the test dataset from using each of the individual learners trained on a specific variable, simply averaging the output from the individual ResNets and using the stacked neural network to combine the outputs. Figures \ref{fig:stacked_z500_5days} and \ref{fig:stacked_t850_5days} show the same comparison for the 5-day prediction error on the test dataset. These figures show that for both the 3-day and 5-day hindcast for both Z500 and T850, the stacked neural network always provides the most accurate result and the error is notably reduced through using it. This is to be expected because the stacked neural network framework (see Figure \ref{fig:schematic}) means that the RMSE achieved from using it is loosely bounded from above by the lowest RMSE achieved by the individual ResNets; otherwise the stacked neural network could achieve a lower error simply by setting that input to 1 and the other inputs to 0. (Note this is only a loose bound because the training dataset of the stacked neural network is not the same as the test dataset on which the predictions are made). Moreover, unlike when we simply average the outputs, the stacked neural network is not adversely influenced by the presence of individual ResNets with much larger errors than others, (for example in Figure \ref{fig:stacked_z500_3days} where the error when using solar radiation is much larger). This is because it learns in training to give a lower weight to these outputs and thus the overall error is unaffected. These accuracy improvements, in addition to the comparatively computationally cheap nature of the stacked neural network (30 minutes on a RTX6000 machine compared to 12 hours to train each individual ResNet on the same machine), show that we are correct to apply the stacked neural network to the ResNet outputs.
	
\begin{figure}[ht]
		\begin{subfigure}{0.49\textwidth}
			\centering
			\includegraphics[height = 0.4\textwidth, width =.9\textwidth]{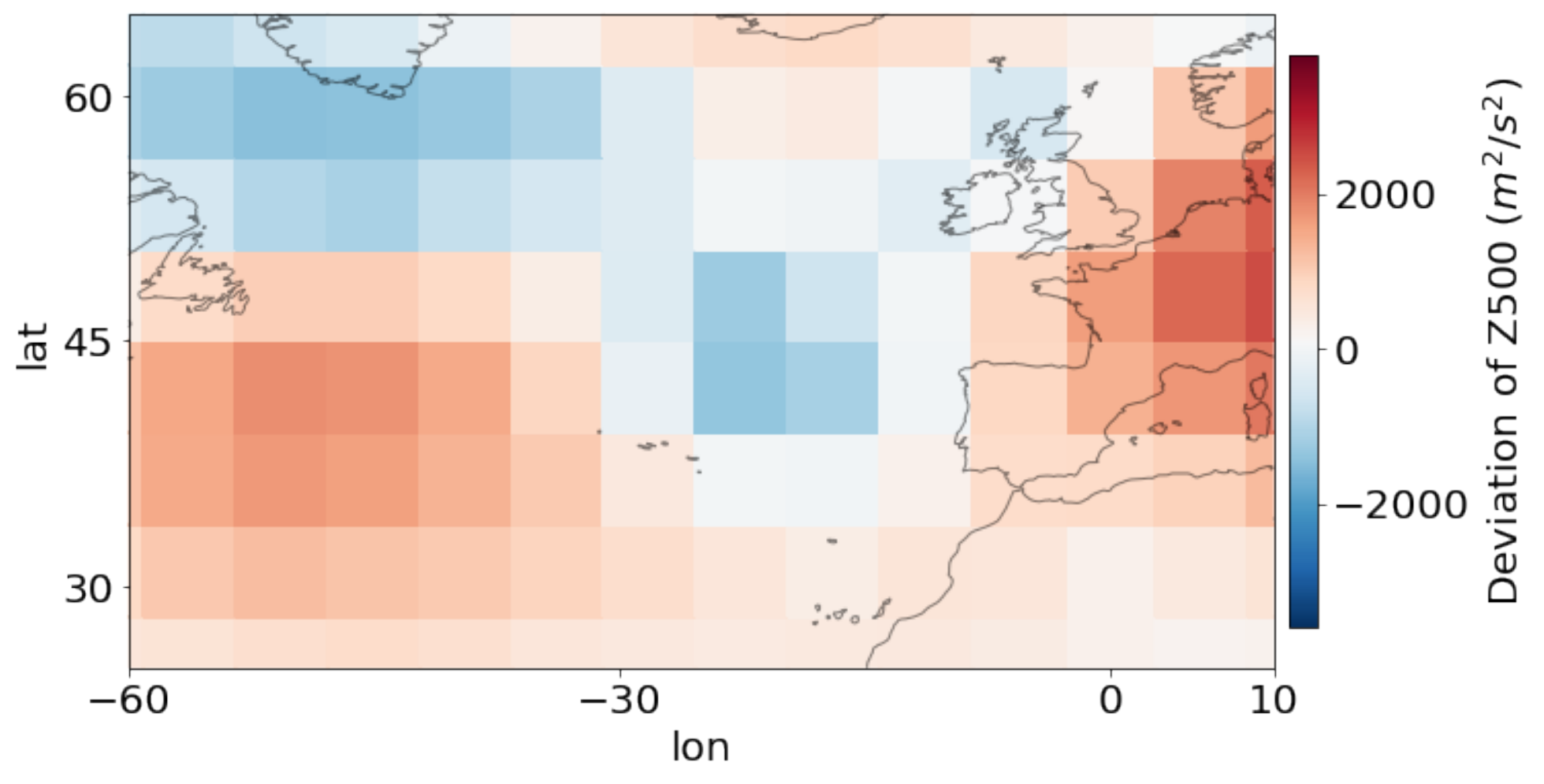}
			\caption{Deviation of true value from climatology - Z500.}
		\end{subfigure}
		\begin{subfigure}{0.49\textwidth}
			\centering
			\includegraphics[height = 0.4\textwidth, width =.9\textwidth]{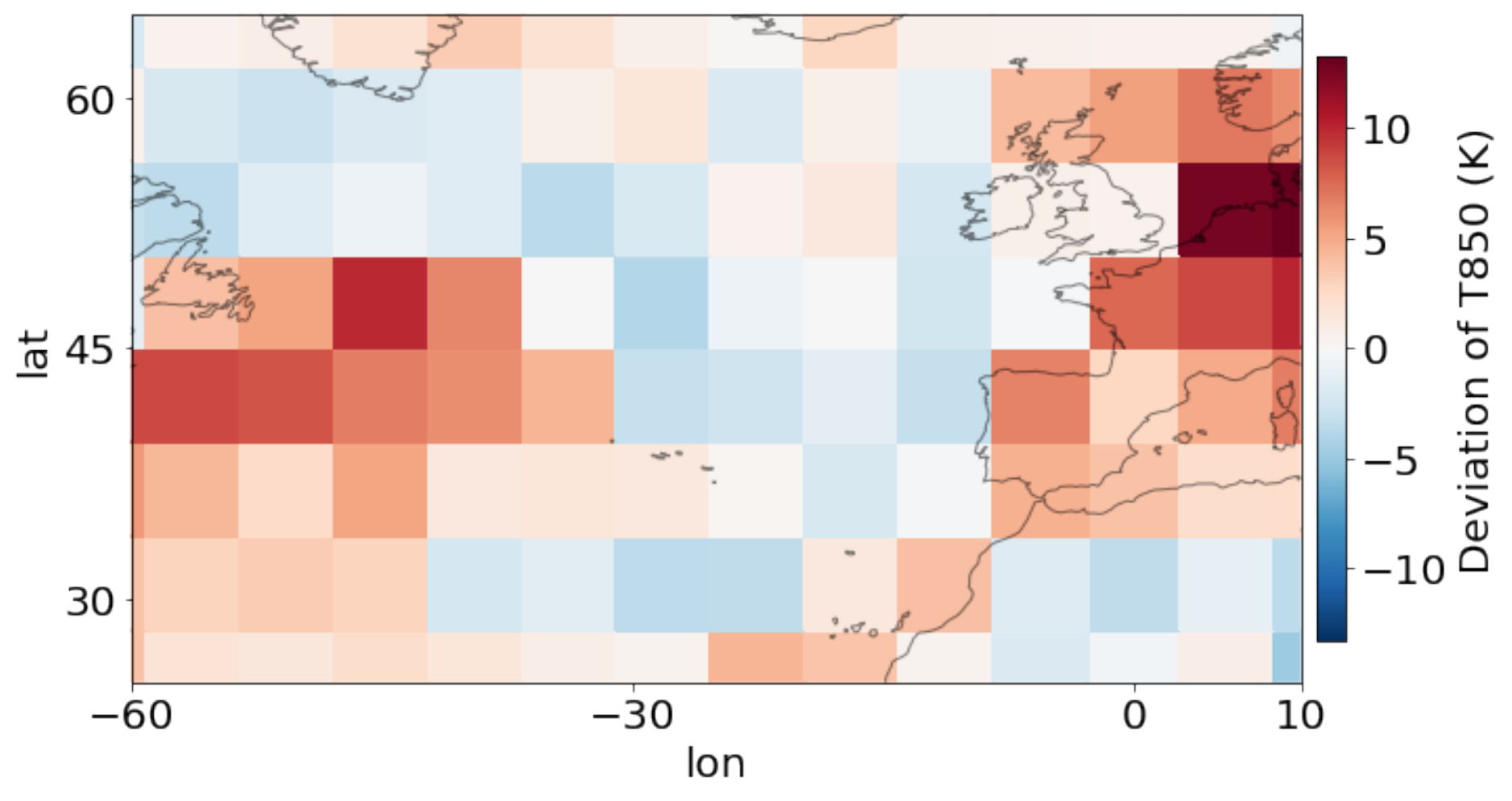}
			\caption{Deviation of true value from climatology - T850.}
		\end{subfigure}
		\begin{subfigure}{0.49\textwidth}
			\centering
			\includegraphics[height = 0.4\textwidth, width =.9\textwidth]{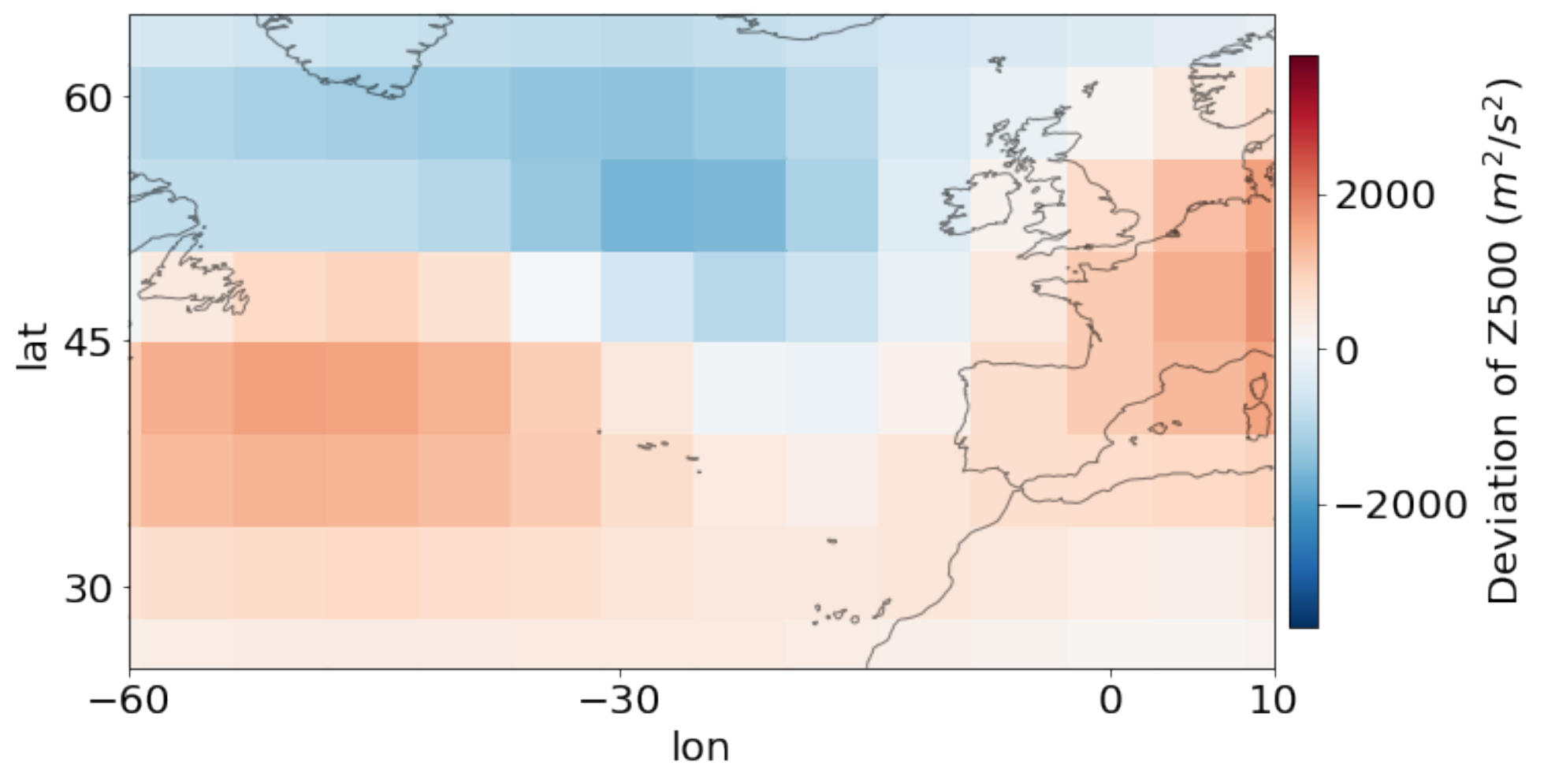}
			\caption{Deviation of 3-day hindcast from climatology -  Z500.}
		\end{subfigure}
		\begin{subfigure}{0.49\textwidth}
			\centering
			\includegraphics[height = 0.4\textwidth, width =.9\textwidth]{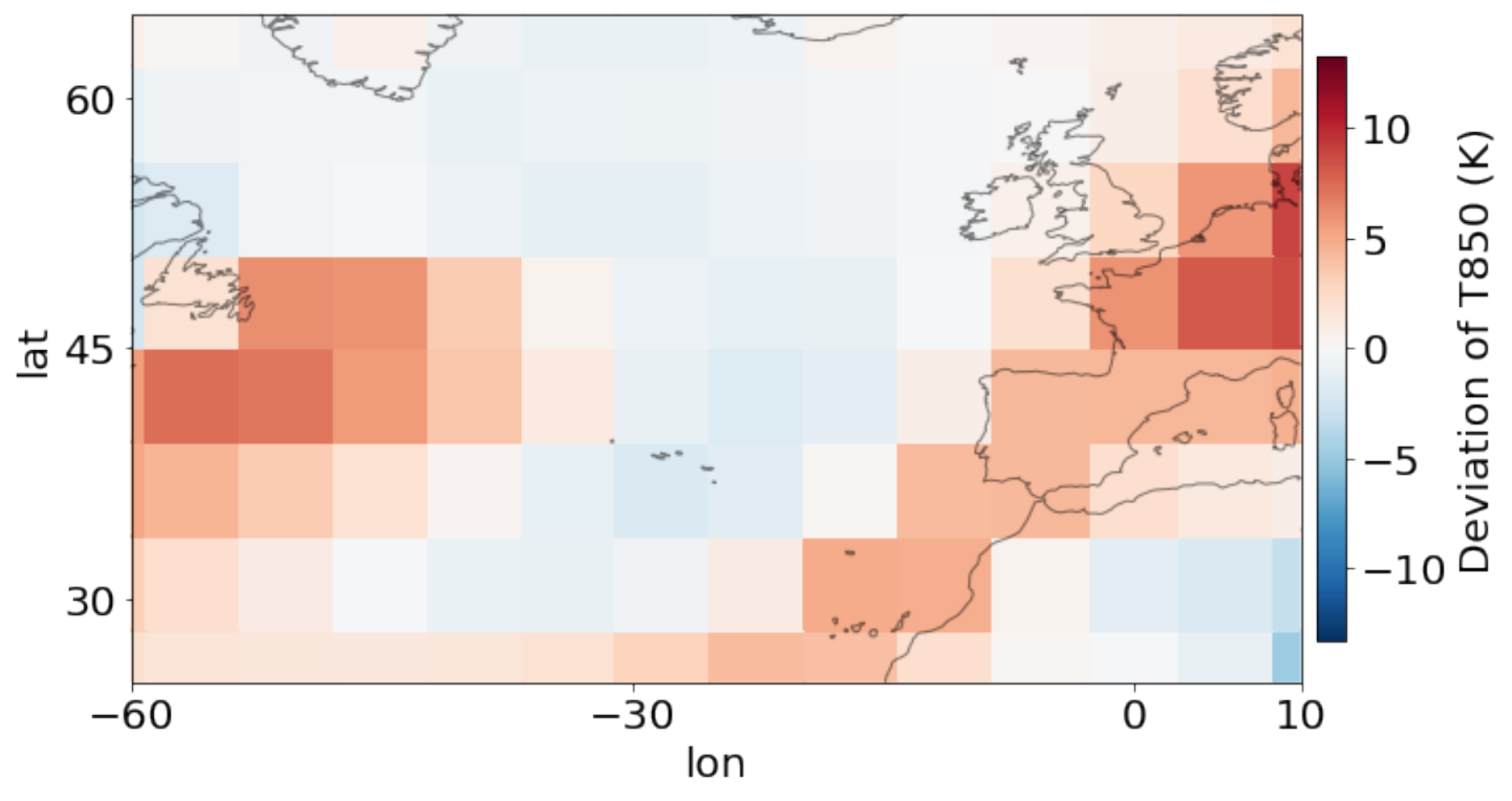}
			\caption{Deviation of 3-day hindcast from climatology -  T850.}
		\end{subfigure}
		\begin{subfigure}{0.49\textwidth}
			\centering
			\includegraphics[height = 0.4\textwidth, width =.9\textwidth]{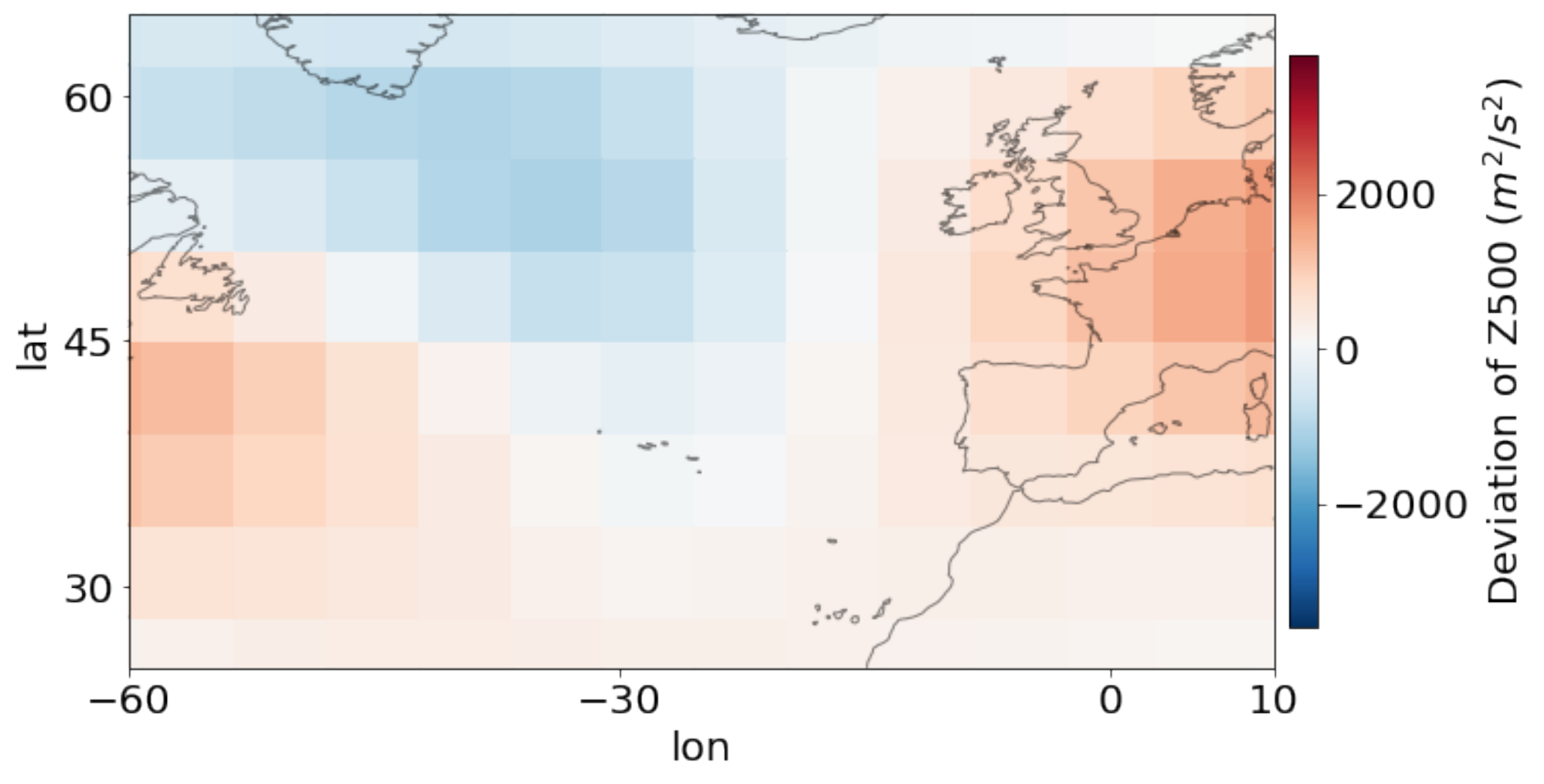}
			\caption{Deviation of 5-day hindcast from climatology -  Z500.}
		\end{subfigure}
		\begin{subfigure}{0.49\textwidth}
			\centering
			\includegraphics[height = 0.4\textwidth, width =.9\textwidth]{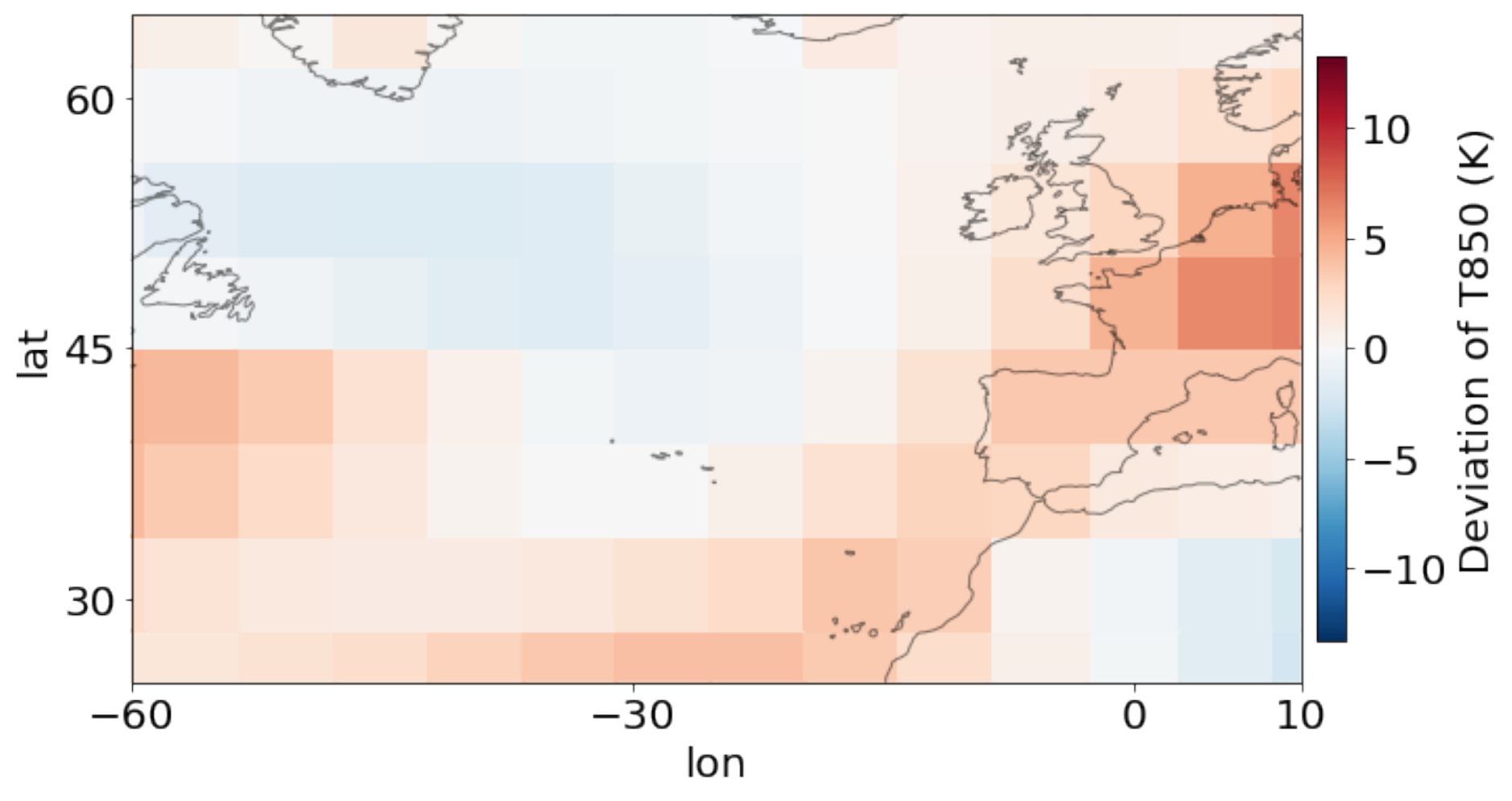}
			\caption{Deviation of 5-day hindcast from climatology -  T850.}
		\end{subfigure}
		\caption{Deviation of the true values, the 3-day hindcast values and the 5-day hindcast values of Z500 and T850, from the climatology value at 00UTC 17 October 2017 during Storm Ophelia.}\label{fig:ophelia}
	\end{figure}

	To visualise what these error values mean, we look at the specific event of Storm Ophelia which was an active storm between 9th-18th October 2017, and the worst storm to affect Ireland in over 50 years \cite[]{storm_ophelia}. We calculate the deviations of the true value, the 3-day and the 5-day hindcasts from the global annual climatology value and show the results in Figure \ref{fig:ophelia} in the North Atlantic region at 00UTC 17 October 2017, which was when the storm was affecting the British Isles. It is clear that for both the 3-day and 5-day hindcast, the neural network can predict the general distribution of Z500 and T850 well. Unsurprisingly, the 3-day hindcast is more accurate at predicting the location of the storm and is able to pick up finer details not present in the 5-day hindcast.
	
	Finally, in Table \ref{final_res_table}, we summarise the results from using our stacked neural network and compare them with results from using simple methods (persistence and climatology), other neural networks \cite[]{Weyn2020,Rasp2020a} and numerical models (IFS T42 and operational IFS) where the numerical model results have been taken from \cite{Rasp2020}. The key finding shown in this table is that our approach is approximately as accurate as that in \cite{Weyn2020} despite the fact that our neural network has a simpler setup than the U-Net approach used in \cite{Weyn2020} and we are using categorical data rather than continuous data which adds an inbuilt error to our results. This highlights the advantages we gain from the data exploration and good choice of neural network architecture described in Section \ref{sec:methods}. The table also shows that our approach is more accurate than the coarse numerical IFS T42 model and the simple methods of persistence and climatology, but less accurate than the neural network approach in \cite{Rasp2020a} and the operational IFS model. It is likely that our neural network's lower skill compared  to \cite{Rasp2020a} is due to the fact that the \cite{Rasp2020a} model is trained on a much larger dataset of the WeatherBench data (117 data variables compared to our training dataset of 34 data variables for Z500 and 33 data variables for T850) and is also pretrained on extra data from the Climate Model Inter-comparison Project (CMIP) \cite[see][]{CMIP}. Thus the approach in \cite{Rasp2020a} is both much more computationally expensive and much more memory intensive than our approach. Furthermore, our approach has introduced improvements which combine dropout based ensembles with the ability to predict probability density functions instead of single values. In the next section, we show how this enables us to make a more informed weather forecast.
	
	\begin{table}[ht]
		\centering
		\begin{tabular}{l||c|c}
			&
			\begin{tabular}[c]{@{}c@{}}\textbf{RMSE Z500 (3-day/5-day)} \\   $(m^{2}/s^{2})$\end{tabular}             & \begin{tabular}[c]{@{}c@{}}\textbf{RMSE T850 (3-day/5-day)} \\ (K)\end{tabular}                  \\ \hline \hline
			\textbf{Stacked neural network}       & \textbf{375/627}                                     & \textbf{2.11/2.91}                                       \\ \hline \hline
			Persistence &  936 / 1033 &
			4.23 / 4.56 \\
			Climatology & 1075 & 5.51 \\
			IFS T42                               &  489 / 743 &  3.09 / 3.83 \\
			\cite{Weyn2020}         &  373 / 611 & 1.98 / 2.87 \\
			\cite{Rasp2020a} & 268 / 499                                                & 1.65 / 2.41                                                \\
			Operational IFS                       &  154 / 334    &  1.36 / 2.03
		\end{tabular}
		\caption[Error for 3-day and 5-day hindcasts]{Error calculated using (\ref{RMSE}) on the test dataset for 3-day and 5-day hindcasts for Z500 and T850. The table compares the results from using our approach in bold, with simple methods (persistence and climatology), numerical models (IFS T42) and other neural networks \cite[]{Weyn2020,Rasp2020a}.}\label{final_res_table}
	\end{table}
	
	\subsection{Estimating uncertainty}\label{sec:uncertainty}
	A common criticism of using neural networks for weather forecasting is that assessing the uncertainty of their forecasts is difficult, and requires techniques such as ensemble approaches, which are often computationally expensive (see \cite[]{Schultz2021} and discussion in Section \ref{sec:intro}). Thus, one of the key novelties of our neural network approach to predicting the weather is that it provides a novel efficient method for producing a probabilistic output, through our prediction of a full probability density function for the variable of interest at each point in space and time.
	This probabilistic output enables us to estimate uncertainty and obtain notably more information from our neural network predictions than can be obtained from a deterministic output.
	
\begin{figure}[ht]
		\centering
		\begin{subfigure}{0.48\textwidth}
			\centering
			\includegraphics[height = 0.6\textwidth, width = 0.85\textwidth, trim={20 3 20 22}, clip]{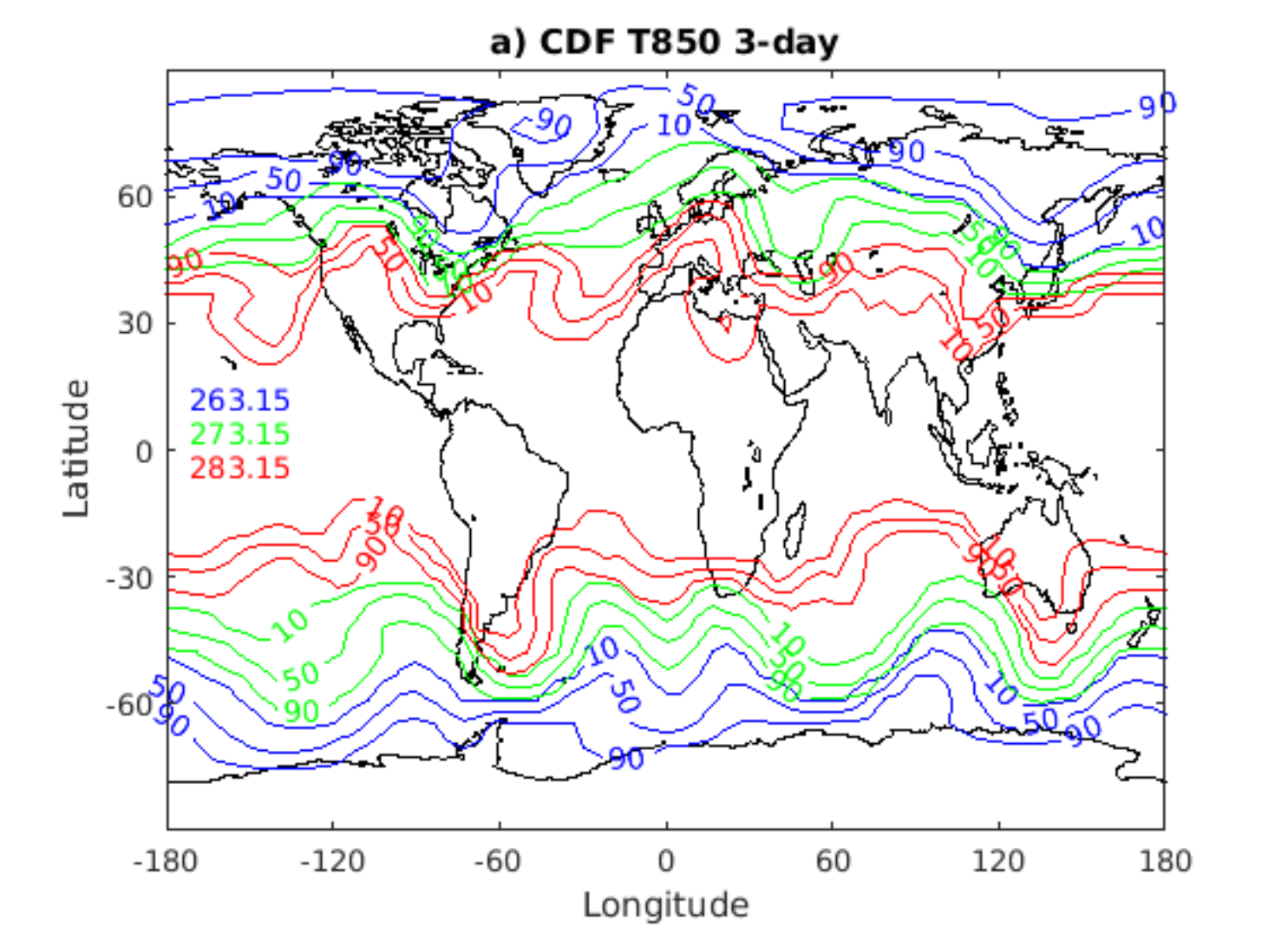}
			\caption{T850 3-day}\label{fig:cdf_maps_t850_3day}
		\end{subfigure}
		\begin{subfigure}{0.48\textwidth}
			\centering
			\includegraphics[height = 0.6\textwidth, width = 0.85\textwidth, trim={20 3 20 22}, clip]{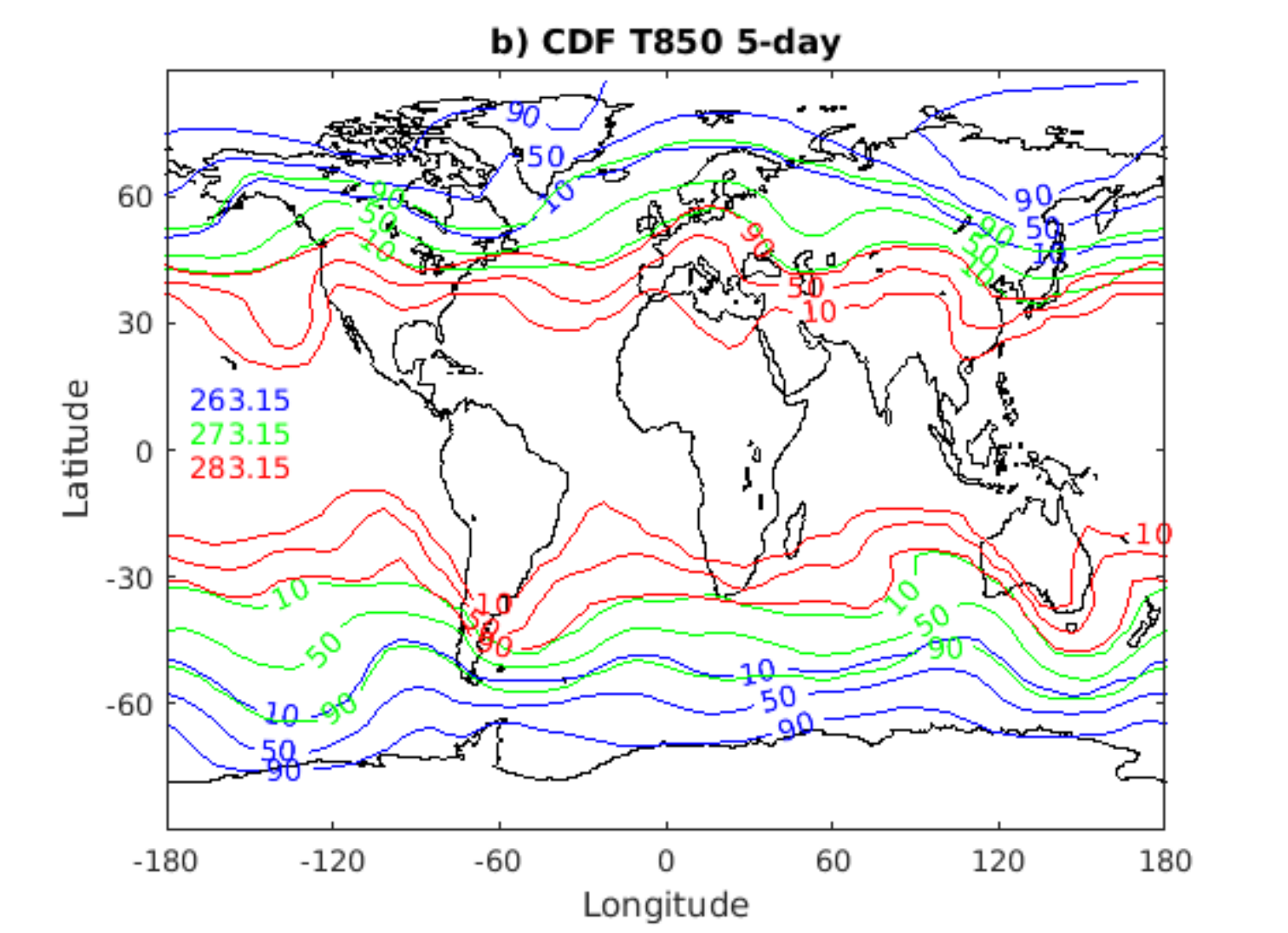}
			\caption{T850 5-day}\label{fig:cdf_maps_t850_5day}
		\end{subfigure}
		\begin{subfigure}{0.48\textwidth}
			\centering
			\includegraphics[height = 0.6\textwidth, width = 0.85\textwidth, trim={20 3 20 22}, clip]{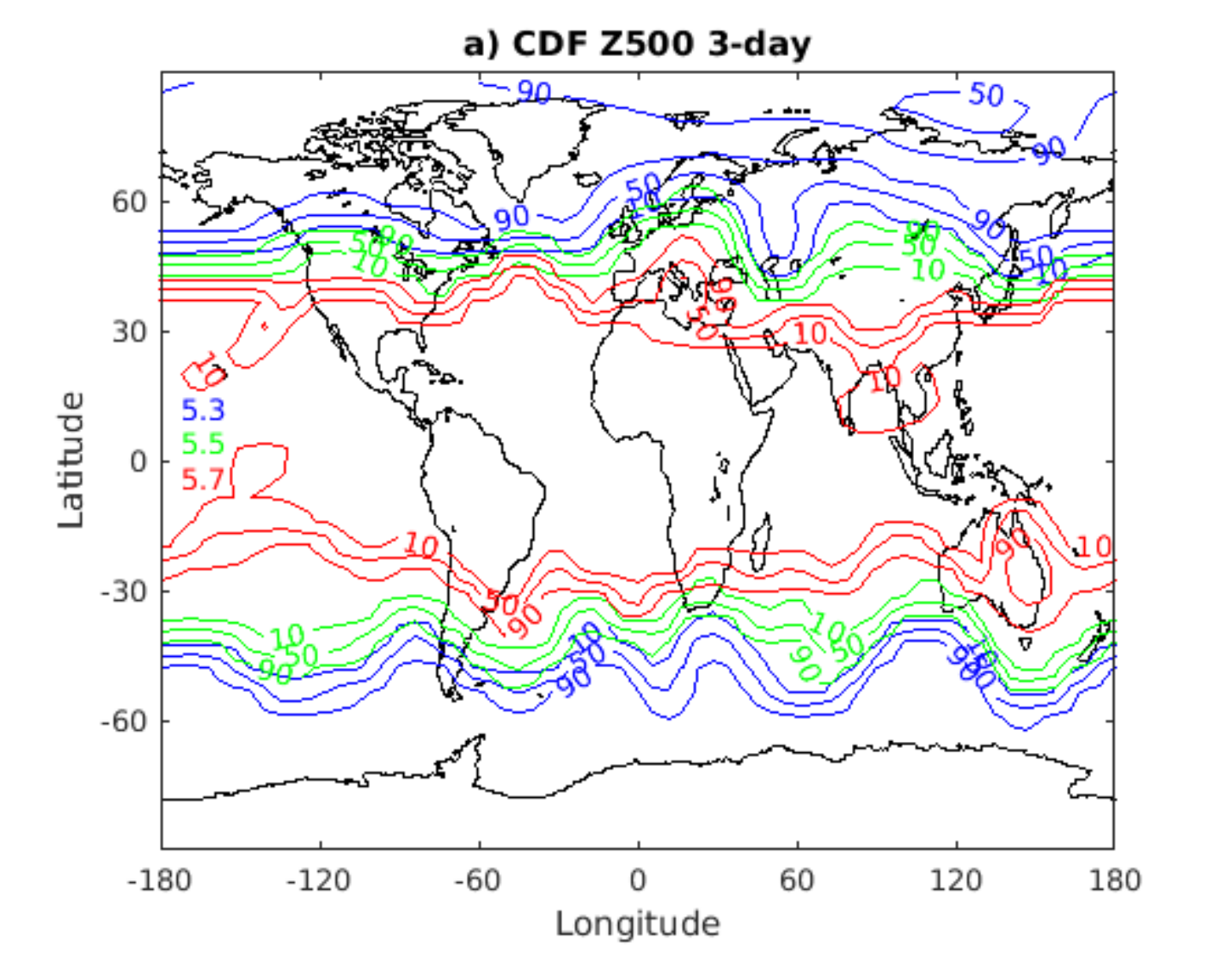}
			\caption{Z500 3-day}\label{fig:cdf_maps_z500_3day}
		\end{subfigure}
		\begin{subfigure}{0.48\textwidth}
			\centering
			\includegraphics[height = 0.6\textwidth, width = 0.85\textwidth, trim={20 3 20 22}, clip]{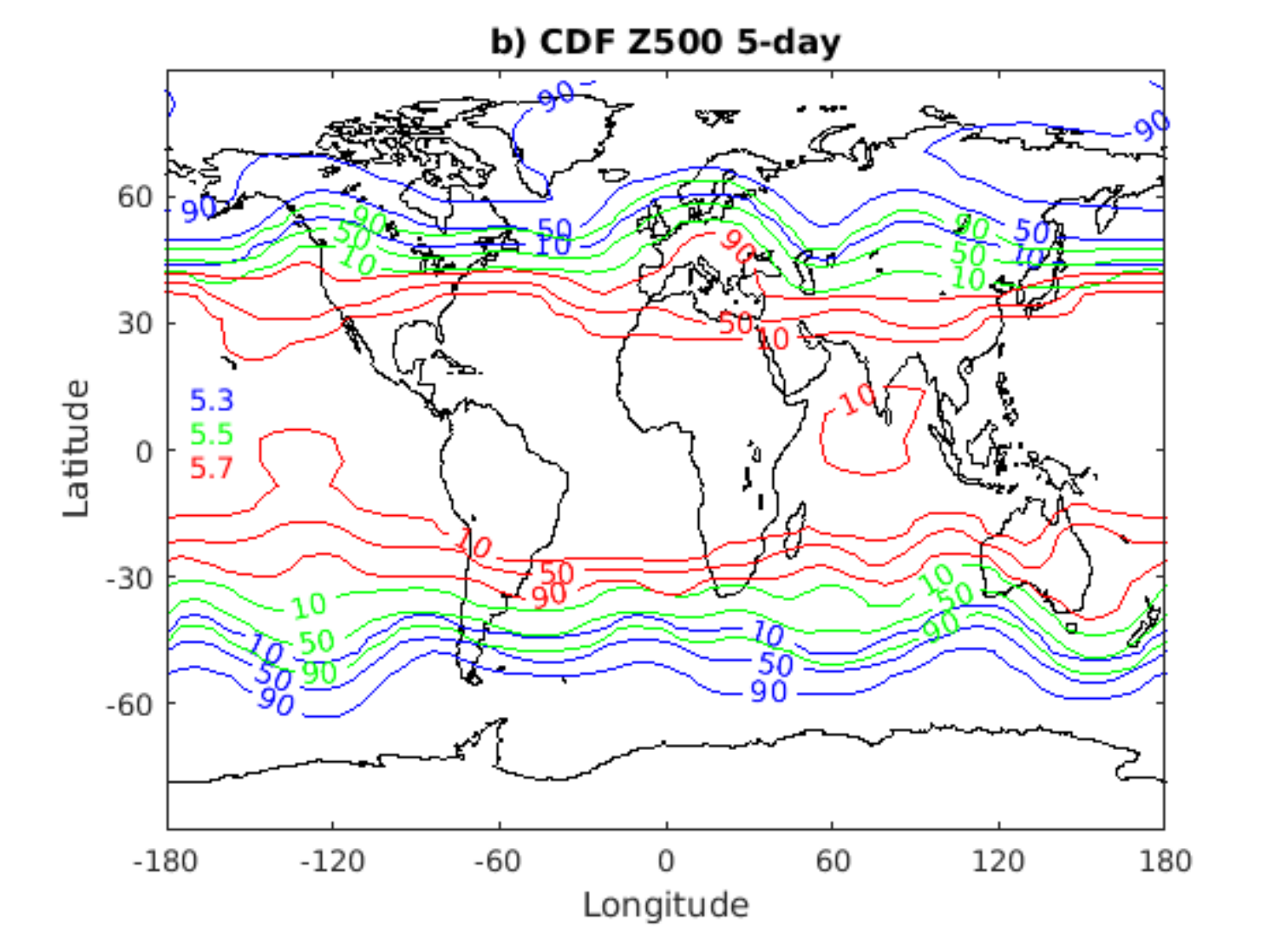}
			\caption{Z500 5-day}\label{fig:cdf_maps_z500_5day}
		\end{subfigure}
		\caption{Contour maps of the cumulative distribution functions at 00UTC 17 Oct 2017. Contours are at 10\%, 50\% and 90\%. For T850 the thresholds considered are T850 < \SI{263.15}{\kelvin} (blue), T850 < \SI{273.15}{\kelvin} (green) and T850 < \SI{283.15}{\kelvin}  (red) and for Z500 they are Z500 < 5.3$\times 10^{4}m$ (blue), Z500 < 5.5$\times 10^{4}m$ (green) and Z500 < 5.7$\times 10^{4}m$ (red).}
	\end{figure}

	To visualise the probabilistic output, we again consider the example of Storm Ophelia. Figures \ref{fig:cdf_maps_t850_3day} and \ref{fig:cdf_maps_t850_5day} show the cumulative distribution function (CDF) for T850 from 3 and 5-day hindcasts at 00UTC 17 Oct 2017. In each panel, the CDFs for three different thresholds are shown using probability contours. These probability contours indicate the locations where the probability of T850 being lower than \SI{263.15}{\kelvin} (-10$^{\circ}C$) (blue), \SI{273.15}{\kelvin} (0$^{\circ}C$) (green) and \SI{283.15}{\kelvin} (10$^{\circ}C$) (red) is 10\%, 50\% and 90\%. Similarly Figures  \ref{fig:cdf_maps_z500_3day} and \ref{fig:cdf_maps_z500_5day} show equivalent probability information for Z500. Note that, although for a given threshold the probability contours will never cross, the probability contours of different thresholds may cross.
	
	In effect, Figures \ref{fig:cdf_maps_t850_3day} and \ref{fig:cdf_maps_t850_5day} shows the median location of the 3 isotherms as well as the 10 to 90$\%$ probability interval. We see meanders in the contours associated with mid-latitude synoptic systems, and locations where the probabilities are not symmetric about the median (e.g. in the 3-day T850 hindcast off the west coast of California, where the distance between the 90\% and 50\% contours is larger than between the 50\% and 10$\%$ contours for the probability of T850 being less than \SI{283.15}{\kelvin} (10$^{\circ}C$)). The CDF for the thresholds of 263.15, 273.15, 283.15 K chosen here, generally do not overlap in the 3-day T850 hindcasts in Figure \ref{fig:cdf_maps_t850_3day}. However, in the 5-day T850 hindcasts, where there is more uncertainty, the spread in each set of contours is wider and there are places where they do overlap (e.g. over Iceland, where there is a 90$\%$ chance of T850 being lower than \SI{273.15}{\kelvin} (0$^{\circ}C$) but also a 10$\%$ chance that it is lower than \SI{263.15}{\kelvin} (-10$^{\circ}C$)). For Z500, Figures \ref{fig:cdf_maps_z500_3day} and \ref{fig:cdf_maps_z500_5day} also show the contours are closer together in the 3-day hindcast compared to the 5-day hindcast, indicating more certainty in the 3-day hindcast.
	
	The probabilistic output of the neural network also means we can estimate uncertainty and skill metrics \citep{hudson2017ensemble} for our predictions. A standard metric used to evaluate probabilistic weather forecasts (see \cite{rasp2018neural}) is the Continuous Ranked Probability Score (CRPS). This metric helps to evaluate the distance between the forecast and the real solution and is calculated using 
	\begin{equation}
	CRPS(F, y) = \int_{-\infty}^{\infty}\left(F(z) - \mathbb{1}(y \leq z)\right)^{2} \,dz
	\end{equation}
	where $F(z)$ is the cumulative distribution function (CDF) of the forecast density function and $\mathbb{1}(y \leq z)^{2}$ is an indicator function representing the CDF of the observed value (see \cite{crps} for more details). Following general practice \cite[e.g.][]{Scher2020}, we compute the CRPS for each density function at each individual gridpoint for each time and then average over them. Note the lower the CRPS score, the better the forecast density function approximates the real density function.
	
	Table \ref{CRPS_tab} shows the CRPS values for Z500 and T850 for the 3 and 5-day hindcasts for our stacked neural network. In order to interpret these values, we first compare the CRPS Z500 values with those obtained in \cite{Scher2020}, who combine the dropout-at-inference ensemble method (see Section \ref{subsubsec:dropout}) with a convolutional neural network to generate an ensemble forecast to predict Z500 in the same time period (2017-2018) as in our work. Our stacked neural network approach performs much better than the approach in \cite{Scher2020} for the 3-day hindcast but much worse for the 5-day hindcast. This suggests that for Z500, the distributions produced by our approach are close to the real distributions for the 3-day hindcast, but suffer greatly as the lead time increases. \cite{Scher2020} did not use their approach to predict T850. Thus instead we compare our T850 values with the `dressed' ERA CRPS values for the T850 5-day hindcast used by ECMWF to benchmark their operational IFS (\cite{Haiden2017} and \cite{Haiden2018}). Note the method to calculate these scores changed between 2017 and 2018, hence the two different values. For reference, we have also included the actual CRPS from the operational IFS \citep{Haiden2018}. Unsurprisingly, the CRPS of the operational IFS is much lower than that from our approach, but the ECMWF benchmark CRPS are fairly close to the CRPS from our approach (especially the interim `dressed' score), which is a promising result. 
	
	\begin{table}[ht]
		\centering
		\begin{tabular}{l|cccc} & \begin{tabular}[c]{@{}c@{}}\textbf{Z500} (3-day)\\ $(m^{2}/s^{2})$\end{tabular} & \begin{tabular}[c]{@{}c@{}}\textbf{Z500} (5-day)\\ $(m^{2}/s^{2})$\end{tabular} & \begin{tabular}[c]{@{}c@{}} \textbf{T850} (3-day)\\ (K)\end{tabular} & \begin{tabular}[c]{@{}c@{}}\textbf{T850} (5-day)\\ (K)\end{tabular} \\ \hline
			\textbf{Stacked neural network}      & 211   & 1500  & 1.22   & 1.69 \\ \hline \hline
			\textbf{\cite{Scher2020}} & 526   & 707    & -     & -   \\
			\textbf{`dressed' ERA interim (2017)}    & -    & -   & -   & 1.44   \\
			\textbf{`dressed' ERA (2018)}   & -   & -       & -    & 1.18   \\
			\textbf{Operational IFS}       & -      & -  & -     & 0.98 
		\end{tabular}
		\vspace{5pt}
		\caption[CRPS]{CRPS for the 3-day and 5-day hindcasts for Z500 and T850 averaged over all gridpoints and time. Where available, the table presents the CRPS values from our stacked neural network, from other neural network approaches \citep{Scher2020}, from the benchmarks used by the ECMWF in 2017 and 2018 and from the Operational IFS for comparison.}\label{CRPS_tab}
	\end{table}
	
	We can further examine the spread of the density functions by looking at their standard deviation, $\sigma$. We do this using 
	\begin{equation}\label{std_dev}
	\sigma = \sqrt{\sum_{i=1}^{100} (x_{i} - \mu)^{2}\mathbb{P}(X = x_{i})},
	\end{equation}
	where $\mu$ is the expectation of X given by Eqn. (ref{expectation}) and, as before, the value of each bin $x_{i}$ is taken to be the lower bound. This informs on the spread of the distribution and allows us to calculate the confidence intervals using 
	\begin{equation}
	\mu \pm  Z \frac{\sigma}{\sqrt{N}},
	\end{equation}
	where N is the number of bins and Z is the z-value taken from the normal distribution and equal to 1.960 for the 95\% confidence interval and 2.576 for the 99\% confidence interval. Note here, the confidence interval is for the distribution at each point in space and time and not for the error as in (\ref{95_conf}) in Section \ref{subsubsec:var_importance}. In Table \ref{table:confidence}, we show the percentage of datapoints in time and space where the true value is within the 95\% confidence interval and 99\% confidence interval around the predicted expected value. The proportions are relatively low but it should be noted that the confidence intervals are relatively narrow because $N$ is relatively large and as shown in Figure \ref{fig:prob_dist}, the distributions are very centred around the expected value -- the average width for the 99\% confidence interval is \SI{0.6}{\kelvin} and \SI{0.8}{\kelvin} for the T850 3-day and 5-day hindcast respectively (smaller than the width of one bin) and \SI{100}{m^{2}s^{-2}} and \SI{160}{m^{2}s^{-2}} for the Z500 3-day and 5-day hindcast respectively (approximately the same size as the width of one bin). Therefore, we also show in Table \ref{table:confidence}, the proportion of true values within one and two $\sigma$ (calculated using (\ref{std_dev})) from the predicted expected value. The majority of true values are within one $\sigma$ of the expected value and almost all (over 90\%) are within two $\sigma$. These metrics, and others like it, help practitioners determine how much confidence to have in the predictions and also allow them to understand the confidence they should have in neural network results compared to other model results.
	
	\begin{table}[ht]
		\centering
		\begin{tabular}{l|cccc}
			& Z500 (3-day) & T850 (3-day) & Z500 (5-day) & T850 (5-day) \\ \hline
			Within 95\% confidence interval & 13.8\%        & 17.2\%        & 14.7\%        & 17.3\%        \\
			Within 99\% confidence interval & 16.3\%        & 20.3\%        & 17.4\%        & 20.5\%        \\
			Within one $\sigma$             & 64.7\%        & 71.2\%        & 67.3\%        & 71.2\%        \\
			Within two $\sigma$             & 93.6\%        & 94.0\%        & 94.0\%        & 94.2\%    \\ 
			\hline
		\end{tabular}
		\vspace{10pt}
		\caption{Percentage of datapoints where the true value is within a set interval from the expected value predicted by the neural network.}\label{table:confidence}
	\end{table}
	
\begin{figure}[ht]
		\begin{subfigure}{0.495\textwidth}
			\centering
			\includegraphics[width = 0.9\textwidth]{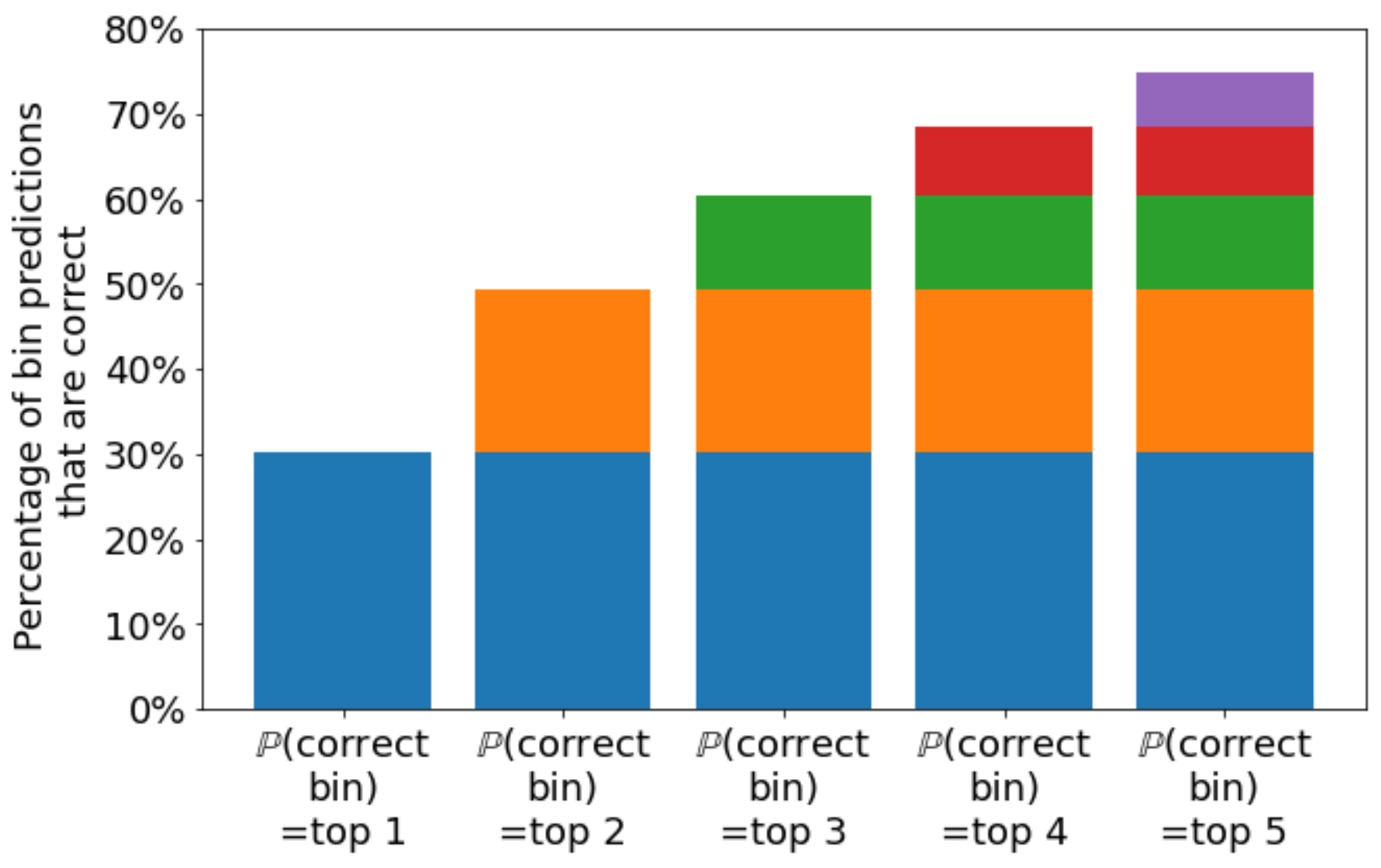}
			\caption{Z500 3-day.}
		\end{subfigure}
		\hfill
		\begin{subfigure}{0.495\textwidth}
			\centering
			\includegraphics[width = 0.9\textwidth]{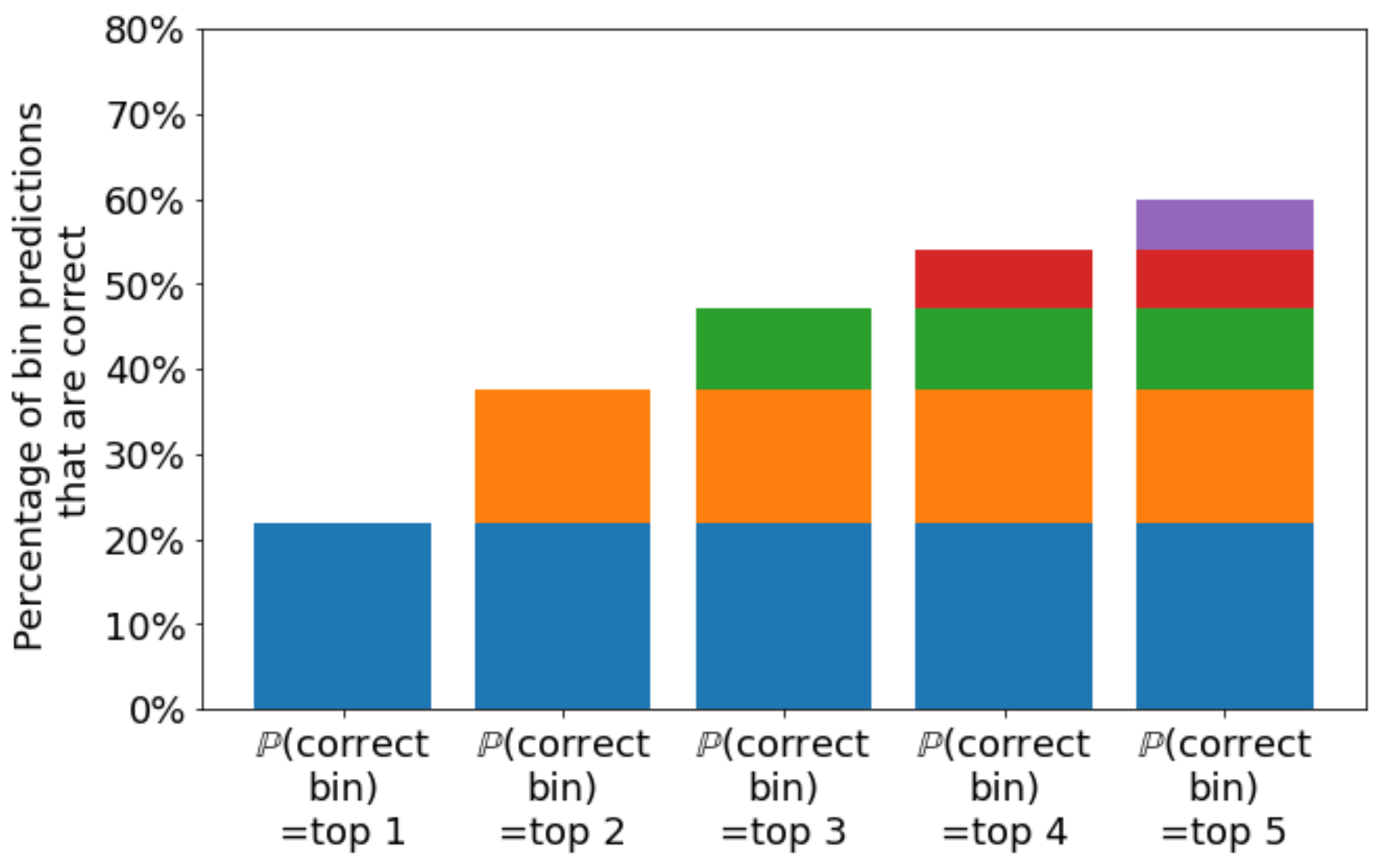}
			\caption{Z500 5-day.}
		\end{subfigure}
		\begin{subfigure}{0.495\textwidth}
			\centering
			\includegraphics[width = 0.9\textwidth]{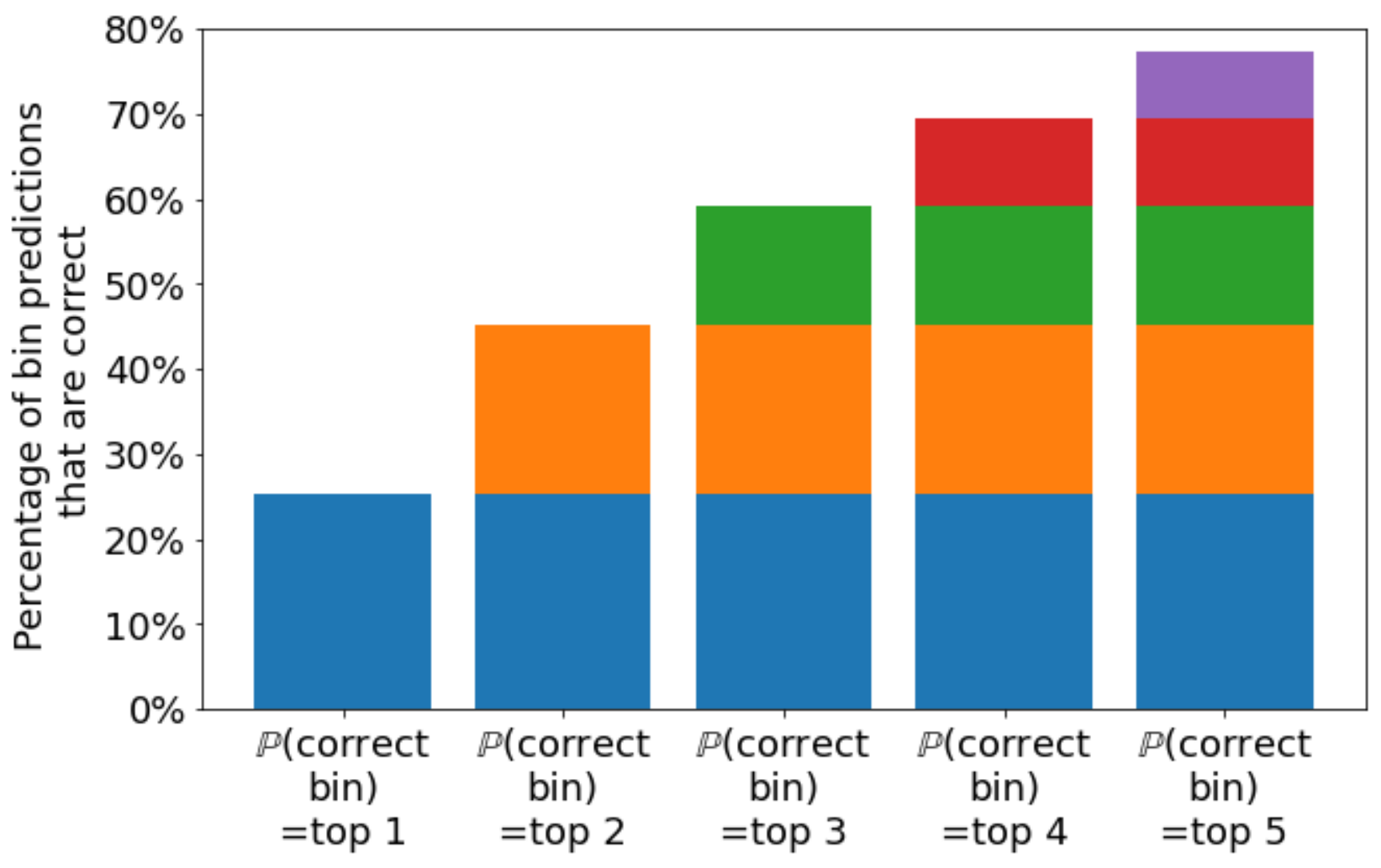}
			\caption{T850 3-day.}
		\end{subfigure}
		\hfill
		\begin{subfigure}{0.495\textwidth}
			\centering
			\includegraphics[width = 0.9\textwidth]{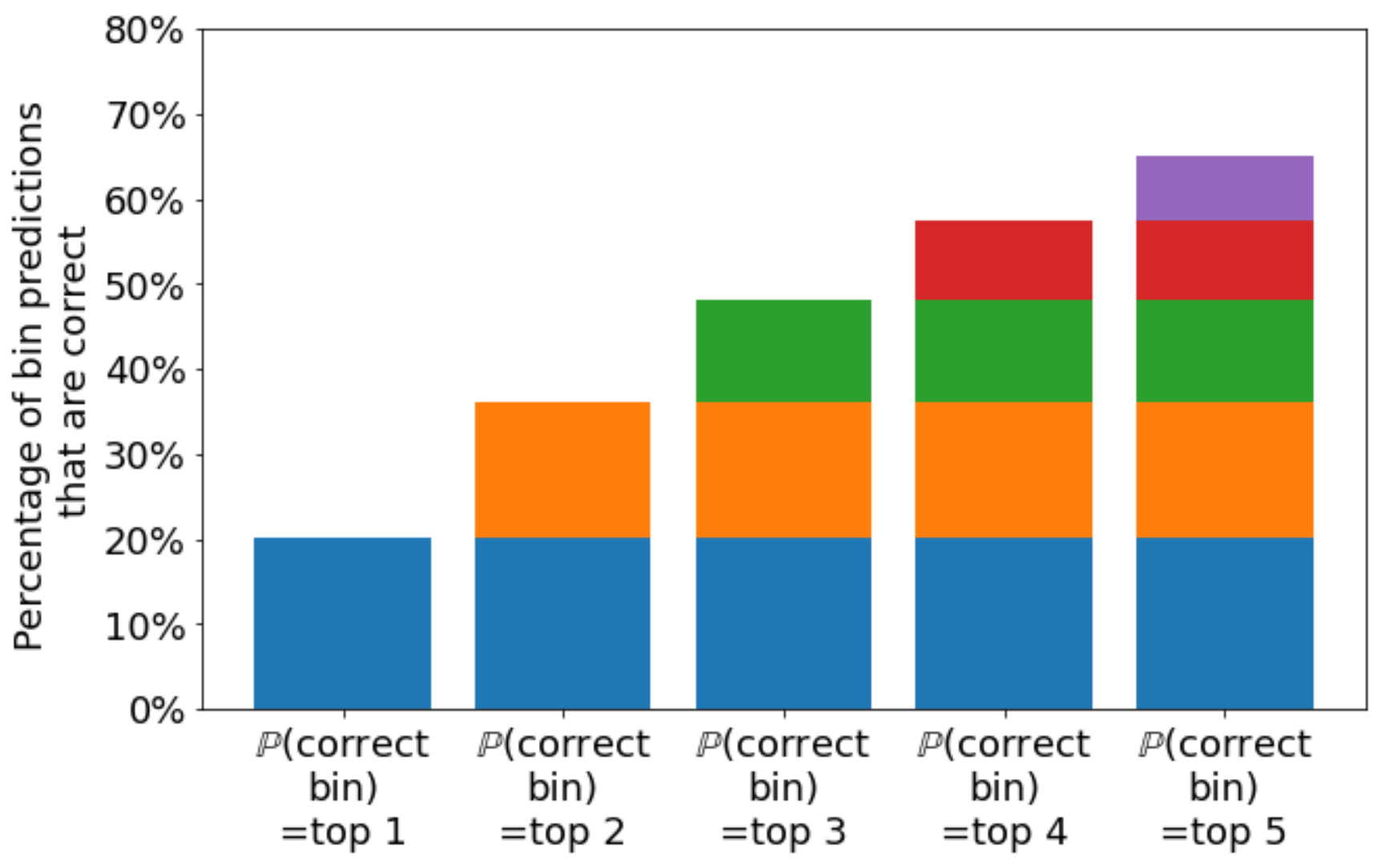}
			\caption{T850 5-day.}
		\end{subfigure}
		\caption{Percentage of all datapoints in time and space where one of the top 1-5 most likely bins predicted by the neural network is the observed bin.}\label{fig:match_prob}
	\end{figure}

	Finally, the probability density functions not only improve understanding of the spread and skill of the results but also inform of scenarios which are not the most likely to occur, but still have a relatively high probability of doing so. Recall from the example in Figure \ref{fig:prob_dist_b}, that in that case the bin the neural network predicts with the highest probability is not the correct bin, but the correct bin still has a high probability of occurring. Figure \ref{fig:match_prob} quantifies cases like these. The first bar in all four subfigures shows the percentage of datapoints for which the bin with the predicted highest probability is the correct bin -- in all four cases this is over 20\%. The second bar shows the percentage of datapoints where the correct bin is either the bin with the predicted highest probability or the predicted second highest probability of occurring. This continues until the fifth bar of the subfigures which shows the percentage of datapoints where the correct bin is one of the top 5 bins that the neural network predicts is most likely to occur. It should be noted here, for clarity, that the SoftMax layer is capable of predicting multi-modal distributions and thus bins with high probabilities are not necessarily close together in value. For example, if the distribution is bimodal, it may be that the bin with the highest probability is at one end of the spectrum and the bin with the second highest probability is at the other end. 
	
	In summary, Figure \ref{fig:match_prob}, shows that for the 3-day predictions, the correct bin is in one of the top 5 most likely to occur for almost 80\% of the datapoints and for the 5-day predictions, the correct bin is one of the top 5 most likely for over 60\% of the datapoints. Given that there are 100 bins such a high proportion from just the top 5 is a notable result. It means that in the vast majority of cases, the true scenarios have a high probability associated with them, which enables practitioners to make informed decisions when forecasting and issuing weather warnings.
	
	Thus we have shown in this section how much more information can be gained for weather forecasting if the neural network predicts density functions rather than single values.
	
	\section{Conclusion}\label{sec:conclusion}
	In this work, we have successfully developed a novel neural network approach, which is able to predict full probability density functions for the target weather variable at each point in space and time instead of single values. This enables practitioners to estimate the uncertainty of the neural network predictions and to provide a more informed weather forecast. In particular, the probability density functions inform about events which, although may not be the most likely to occur, still have a significant probability of happening. We have thus provided a strong proof-of-concept of how neural networks can be used to produce probabilistic weather forecasts, which is an area where many weather forecasting practitioners would like to see neural networks improve \cite[see][]{Schultz2021}.   
	
	For our neural network predictions, a relatively simple ResNet has been used and carefully optimised to improve the accuracy of our results. Through the careful choice of this architecture along with extensive data exploration, we have produced weather hindcasts which are more accurate than some coarse NWP models, and as accurate as those in \cite{Weyn2020} which uses a more complex neural network architecture. We have also shown that the important variables identified by the data exploration agree with physical reasoning, thereby validating our neural network approach. Moreover, our novel use of a stacked neural network to combine outputs for weather forecasting reduces the memory cost, as well as the computational cost as smaller networks generally take less time to train, and means that less powerful computers are required to make predictions.  
	
	Finally, in this work we have shown that transforming our output data to categorical can still give accurate results for continuous numerical weather data. We have shown that it is possible to move beyond point estimates for neural networks based weather forecasts and produce a probabilistic forecast by combining multiple smaller more efficient models. This opens up a new avenue of research for data-driven weather forecasting, where ever bigger models trained with enormous datasets is not the only way to achieve model skill. In future work, we will seek to explore the use of transfer learning to further improve the training efficiency of the neural networks.
	
	\subsection{Computer Code availability}\label{code}
	The relevant code for the neural networks presented in this work can be found at \url{https://github.com/mc4117/ResNet_Weather.git}.
	
	\subsection*{Acknowldgements}
	We thank Gabriele Messori and an anonymous reviewer for their comments which helped improve the manuscript.
	
\bibliographystyle{apalike}

\bibliography{References.bib}

\end{document}